\documentclass[10pt,twocolumn,letterpaper]{article}

\usepackage{cvpr}

\usepackage{graphicx}
\usepackage{amsmath}
\usepackage{amssymb}
\usepackage{booktabs}
\usepackage{multirow}
\usepackage{colortbl}
\usepackage{enumitem}
\usepackage{xspace}
\usepackage{paralist}
\usepackage{adjustbox}
\usepackage{multibib}
\usepackage[dvipsnames]{xcolor}
\usepackage{microtype}

\newcommand{\figref}[1]{Fig.~\ref{#1}}
\newcommand{\tabref}[1]{Tab.~\ref{#1}}
\newcommand{\secref}[1]{Sec.~\ref{#1}}

\newcommand{\equref}[1]{Eq.~(\ref{#1})}

\definecolor{rred}{RGB}{245, 152, 153}
\definecolor{oorange}{RGB}{253, 205, 154}
\definecolor{yyellow}{RGB}{248,244,140}

\newcommand{\Ra}[1]{{\textbf{\color{magenta}{R1}}}}
\newcommand{\Rb}[1]{{\textbf{\color{blue}{R2}}}}
\newcommand{\Rc}[1]{{\textbf{\color{cyan}{R3}}}}
\newcommand{\methodname}{Vid2Avatar-Pro\xspace}
\newcommand{\suppmat}{Supp.~Mat\xspace}

\DeclareMathOperator{\diag}{diag}

\definecolor{cvprblue}{rgb}{0.21,0.49,0.74}
\usepackage[pagebackref,breaklinks,colorlinks,allcolors=cvprblue]{hyperref}
\usepackage[para]{footmisc}

\title{Vid2Avatar-Pro: Authentic Avatar from Videos in the Wild via Universal Prior}

\author{Chen Guo$^{*1,2}$ \quad Junxuan Li$^{*1}$ \quad Yash Kant$^{1,3}$ \quad Yaser Sheikh$^{1}$ \quad Shunsuke Saito$^{\dag1}$ \quad Chen Cao$^{\dag1}$ \\
 $^1$Codec Avatars Lab, Meta \quad 
 $^2$ETH Z{\"u}rich \quad
 $^3$University of Toronto
 \\
}

\begin{document}

\twocolumn[{%
\renewcommand\twocolumn[1][]{#1}%
\maketitle
% \vspace{-3.5em}
\begin{center}
    \captionsetup{type=figure}   \includegraphics[width=\linewidth,trim=0 0 0 0,clip]{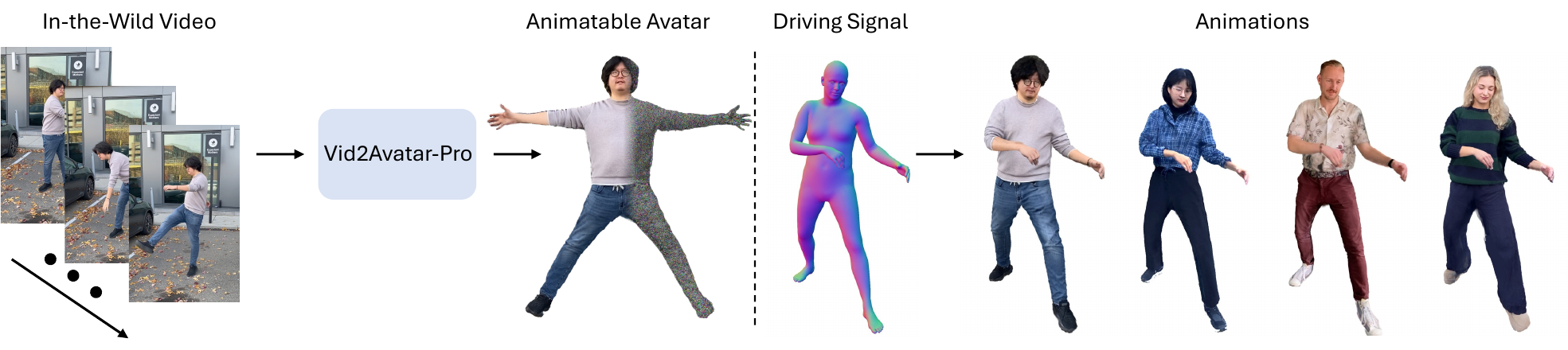}
    \caption{We present \methodname, a method to create photorealistic 3D human avatars from monocular in-the-wild videos via pre-trained universal prior. Our method can faithfully create high-fidelity human avatars from a single video and generate realistic animations.}
    \label{fig:teaser}
\end{center}%
}]

%
% Reduce space between the footnote marker and the text
\makeatletter
\renewcommand\@makefntext[1]{%
  \setlength{\parindent}{0pt} % Remove indent
  \@makefnmark #1%
}
\makeatother
\def\thefootnote{*}\footnotetext[1]{Equal contribution}%
\def\thefootnote{\dag}\footnotetext[1]{Equal advisory}

\newcommand{\figureteaser}{

\begin{figure*}[t]
% \raggedleft
\includegraphics[width=\linewidth,trim=0 5 0 0,clip]{figures/teaser.pdf}
% \fbox{\rule{0pt}{2in} \rule{.9\linewidth}{0pt}}

\caption{\textbf{Teaser Placeholder.} .}

\label{fig:teaser}
\end{figure*}
}

\newcommand{\figurePipeline}{

\begin{figure*}[t]
\includegraphics[width=\linewidth, trim=0 10 0 0,clip]{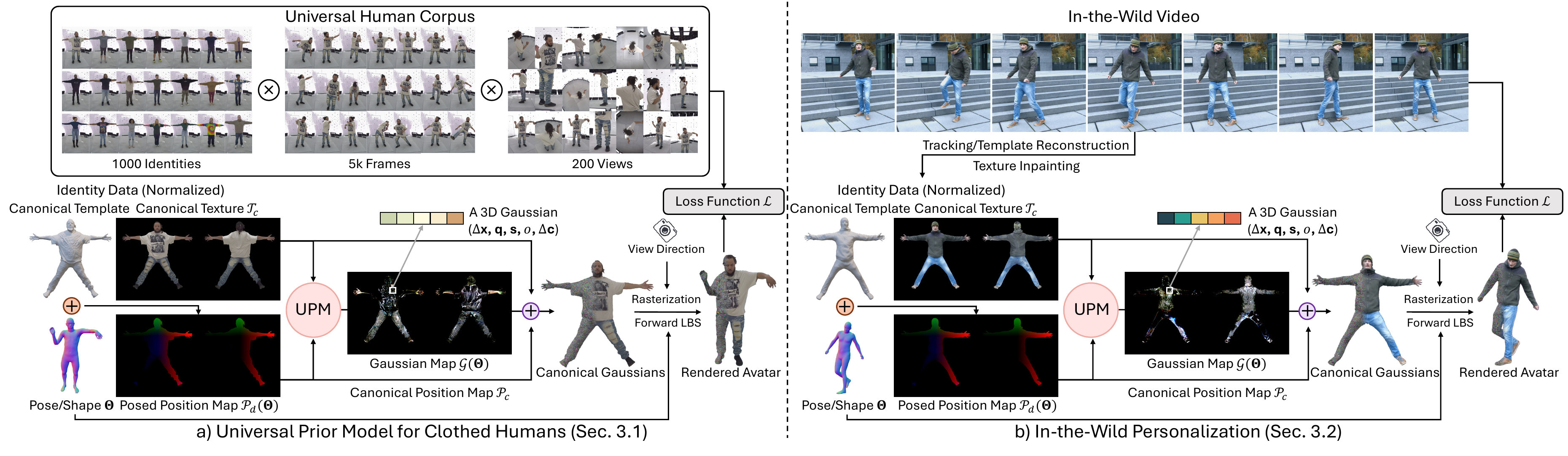}
% \fbox{\rule{0pt}{2in} \rule{.9\linewidth}{0pt}}
\caption{\textbf{Method Overview.} a) We employ a large corpus of multi-view dynamic clothed human performances to train a cross-identity universal prior model (UPM). During training, UPM is conditioned on the normalized identity-specific texture map $\mathcal{T}_c$ and takes the posed position map $\mathcal{P}_d(\boldsymbol{\Theta})$ as input to predict Gaussian attributes. We extract the canonical 3D Gaussians and synthesize human rendering for training pose/shape parameters $\boldsymbol{\Theta}$ by applying forward LBS and rasterization. We minimize the loss $\mathcal{L}$ over the entire universal human corpus. 
b) Given a monocular in-the-wild video of an unseen identity, we track the human pose/shape parameters $\boldsymbol{\Theta}$ and reconstruct the canonical textured template. We further deploy a diffusion-based model tailored for canonical texture inpainting to complete the canonical texture map. We then fine-tune our pre-trained UPM on the monocular observations via inverse rendering to recover person-specific details.}
\label{fig:pipeline}
\end{figure*}
}

\newcommand{\figureneuman}{

\begin{figure*}[t]
% \raggedleft
\includegraphics[width=\linewidth,trim=0 10 0 0,clip]{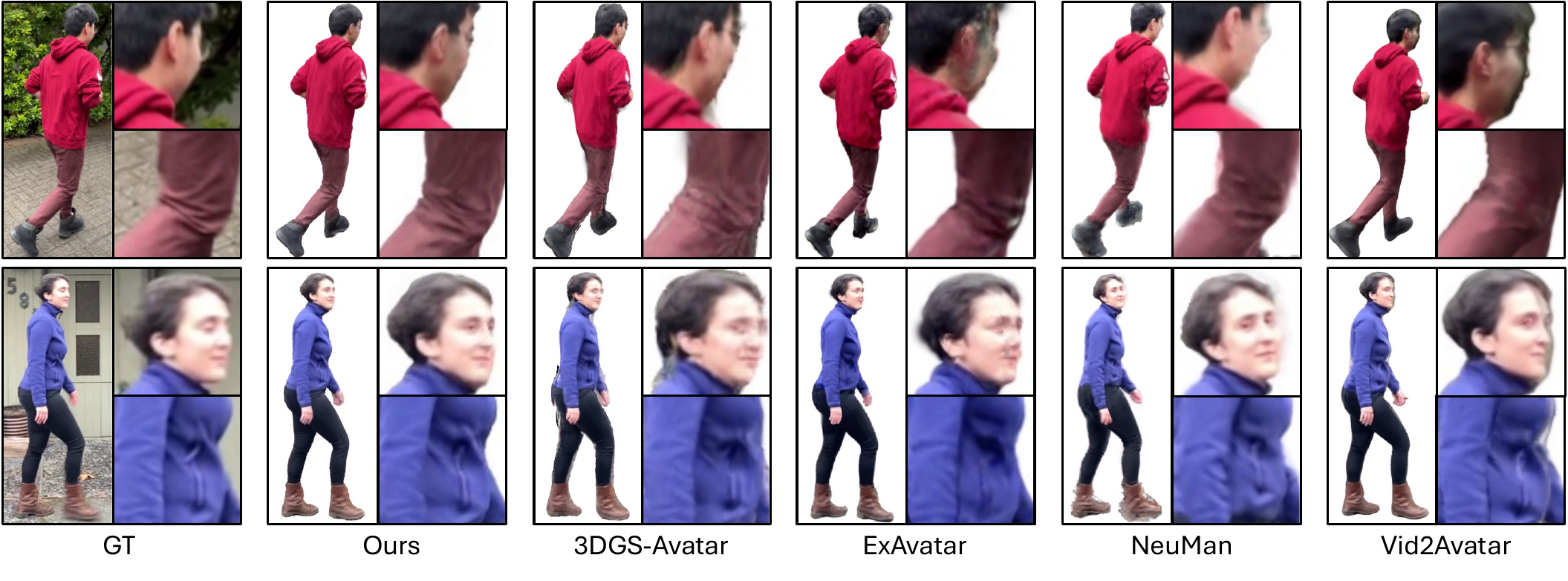}
% \fbox{\rule{0pt}{2in} \rule{.9\linewidth}{0pt}}

\caption{\textbf{Qualitative interpolation synthesis comparisons on NeuMan dataset.} Baseline methods tend to render with artifacts (\eg, corrupted faces and feet) and less details (\eg, the clothing wrinkles and the clothes zipper). In contrast, our method generates clean and realistic human renderings while recovering more appearance details (\eg, facial features).}

\label{fig:neuman}
\end{figure*}
}

\newcommand{\figuremonoperfcap}{

\begin{figure*}[t]
% \raggedleft
\includegraphics[width=\linewidth,trim=0 10 0 0,clip]{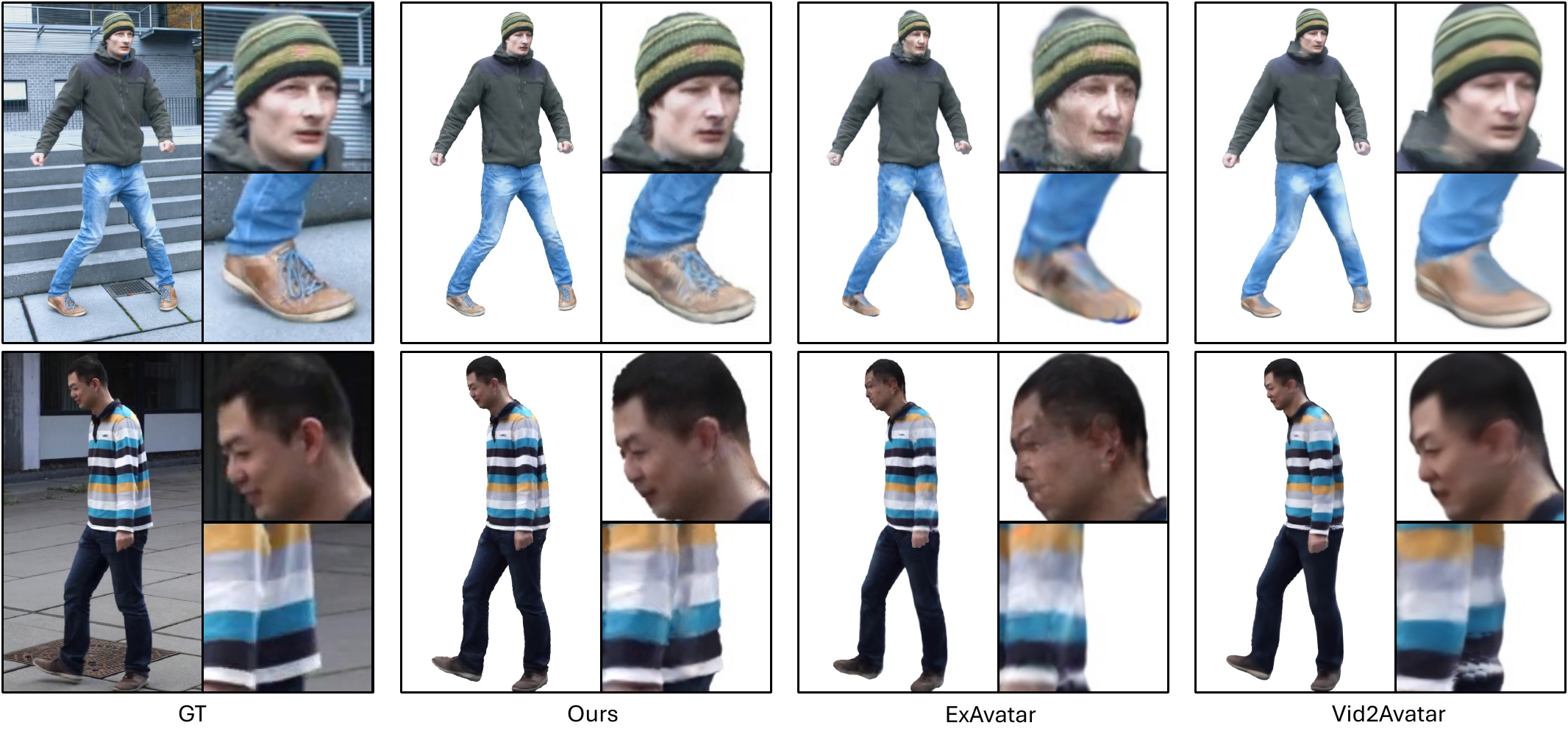}
% \fbox{\rule{0pt}{2in} \rule{.9\linewidth}{0pt}}

\caption{\textbf{Qualitative extrapolation synthesis comparisons on MonoPerfCap dataset.} In this more challenging setting, our method produces more identity-preserving human renderings with finer-grained appearance details, \eg, the facial features and shoes. Powered by our universal prior model, \methodname generates more plausible pose-dependent surface deformations (\cf the hem of the T-shirt).}

\label{fig:monoperfcap}
\end{figure*}
}

\newcommand{\figureavgscale}{

\begin{figure}[t]
% \raggedleft
\includegraphics[width=\linewidth,trim=0 10 0 0,clip]{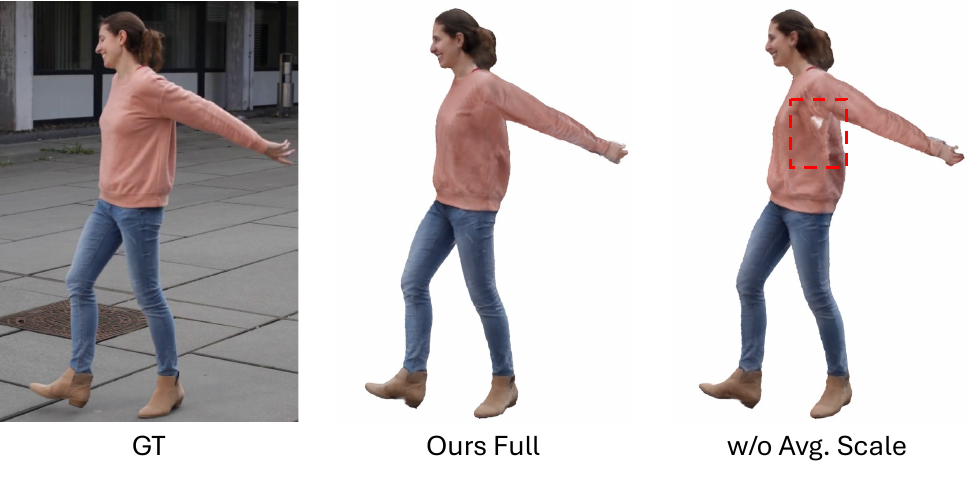}
% \fbox{\rule{0pt}{2in} \rule{.9\linewidth}{0pt}}
\caption{\textbf{Importance of skeleton-based normalization.} Without skeleton-based conditioning data normalization, the created avatar tends to produce artifacts when driven with out-of-distribution poses (\eg, the holes close to the armpit).}

\label{fig:ablation_avg_scale}
\end{figure}
}

\newcommand{\figureinpainting}{

\begin{figure}[t]
% \raggedleft
\includegraphics[width=\linewidth,trim=0 10 0 0,clip]{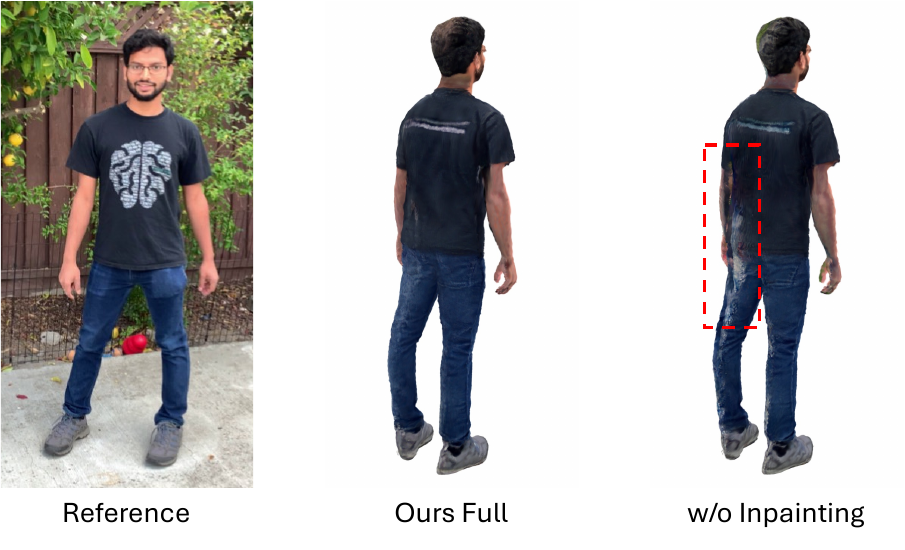}
% \fbox{\rule{0pt}{2in} \rule{.9\linewidth}{0pt}}
\caption{\textbf{Importance of inpainting.} Our diffusion-based inpainting module can effectively complete the textures that are missing from the monocular observations.}

\label{fig:ablation_inpainting}
\end{figure}
}

\newcommand{\figurefinetune}{

\begin{figure}[t]
% \raggedleft
\includegraphics[width=\linewidth,trim=0 5 0 0,clip]{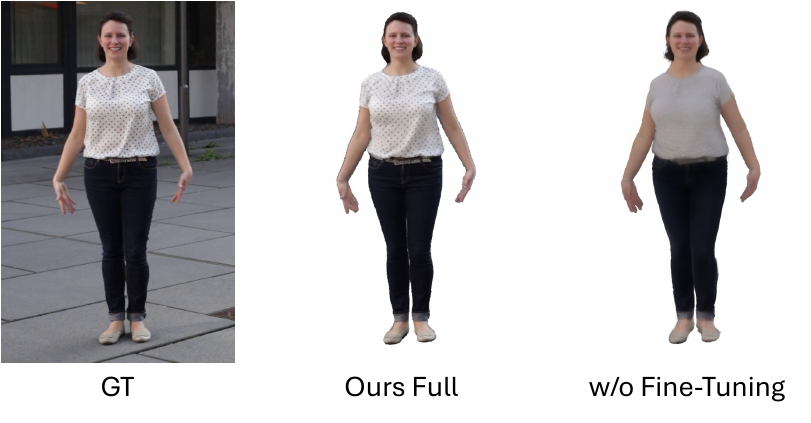}
% \fbox{\rule{0pt}{2in} \rule{.9\linewidth}{0pt}}
\caption{\textbf{Importance of fine-tuning.} Without fine-tuning, the human avatar cannot preserve the fine-grained human appearance details such as the T-shirt pattern and the belt.}

\label{fig:ablation_finetune}
\end{figure}
}

\newcommand{\figurequali}{

\begin{figure*}[t]
% \raggedleft
\includegraphics[width=\linewidth,trim=0 10 0 0,clip]{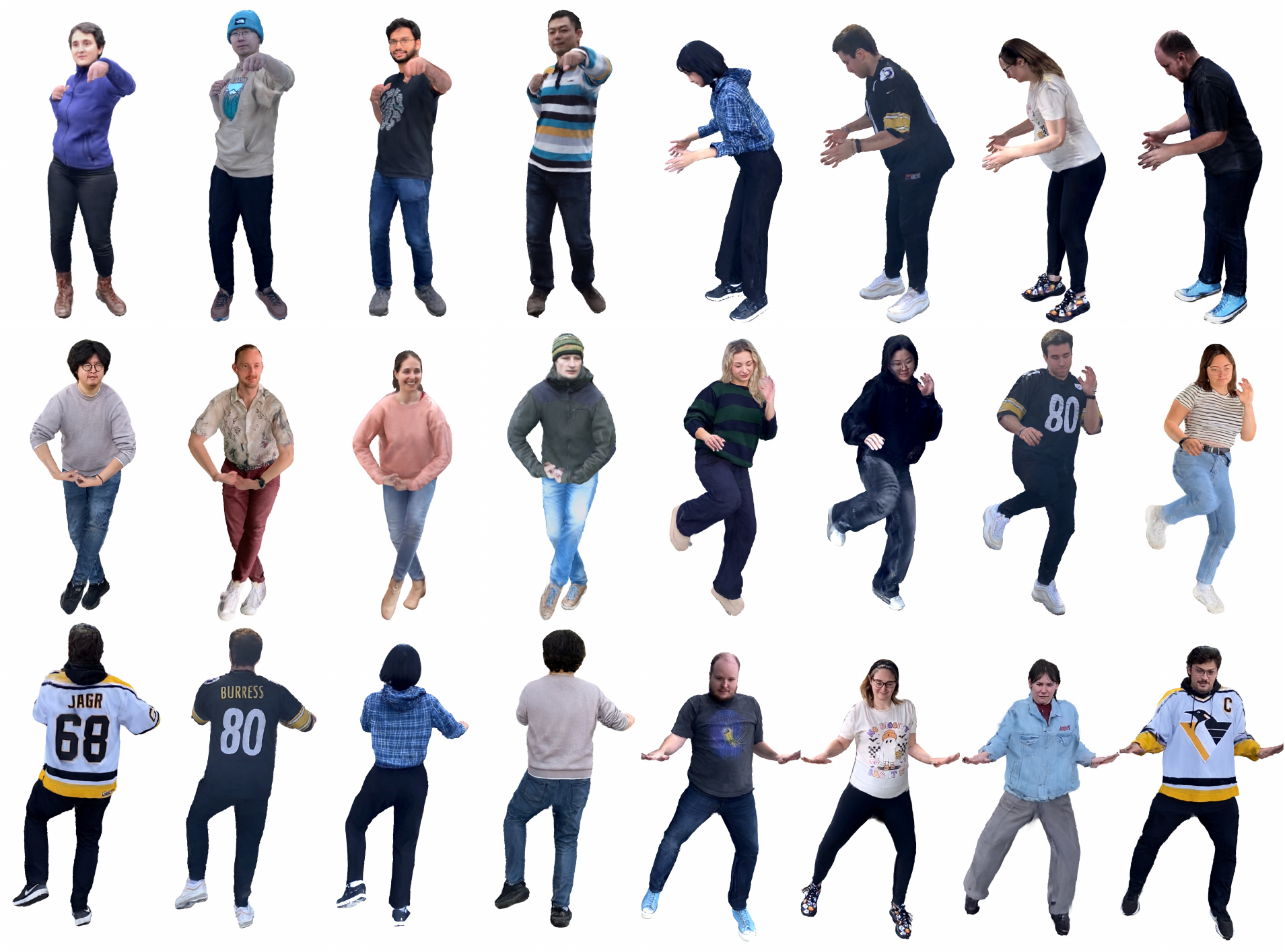}
% \fbox{\rule{0pt}{2in} \rule{.9\linewidth}{0pt}}

\caption{\textbf{Visual animation results of avatars created from monocular in-the-wild videos.} The created 3D avatars can be animated using novel human poses and demonstrate highly detailed appearance from arbitrary view points.}
\label{fig:quali_res}
\end{figure*}
}

\newcommand{\figurenet}{

\begin{figure}[t]
\raggedleft
\includegraphics[width=\linewidth,trim=0 0 0 0,clip]{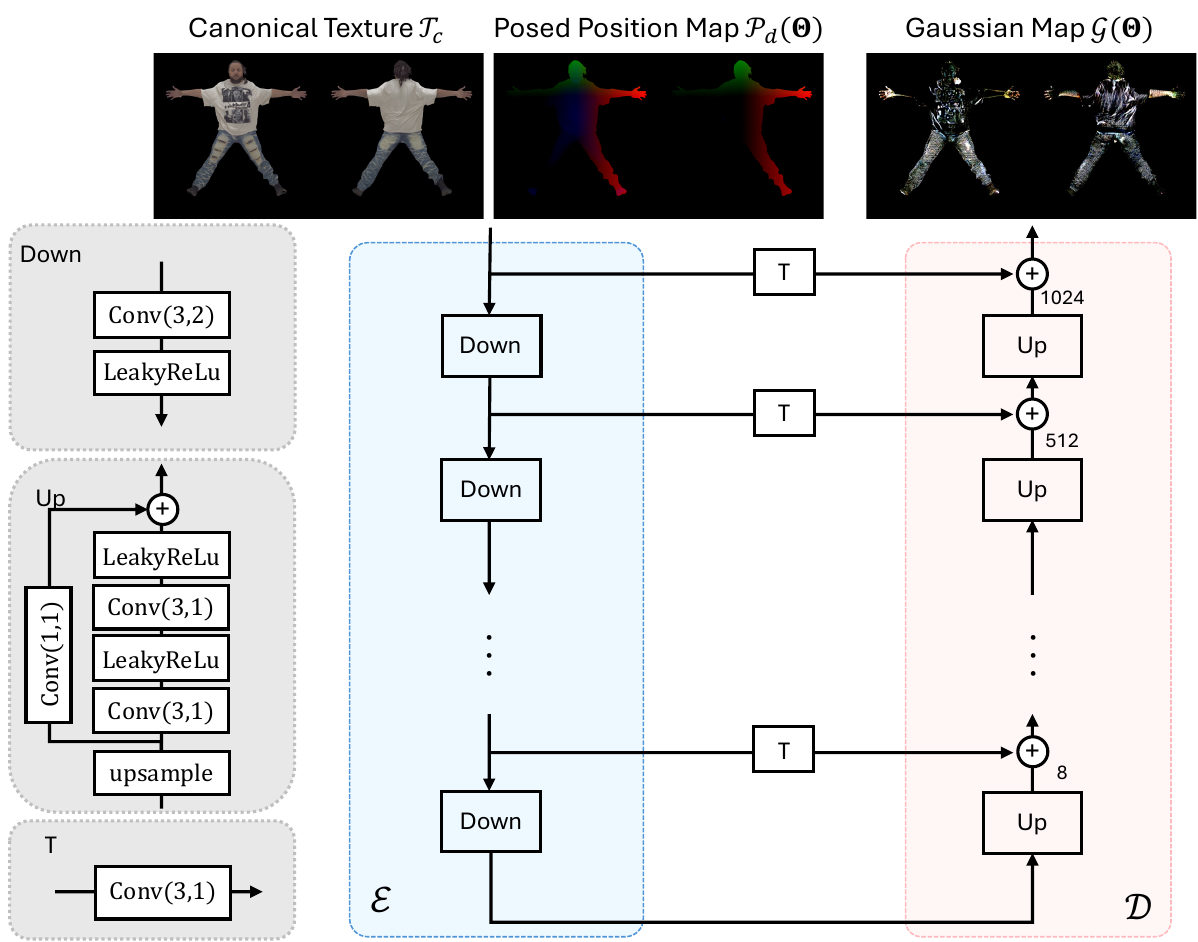}
% \fbox{\rule{0pt}{2in} \rule{.9\linewidth}{0pt}}

\caption{\textbf{Universal Prior Model Architecture.}}
\label{fig:net}
\end{figure}
}

\newcommand{\figurecanotex}{

\begin{figure}[t]
\raggedleft
\includegraphics[width=\linewidth,trim=0 0 0 0,clip]{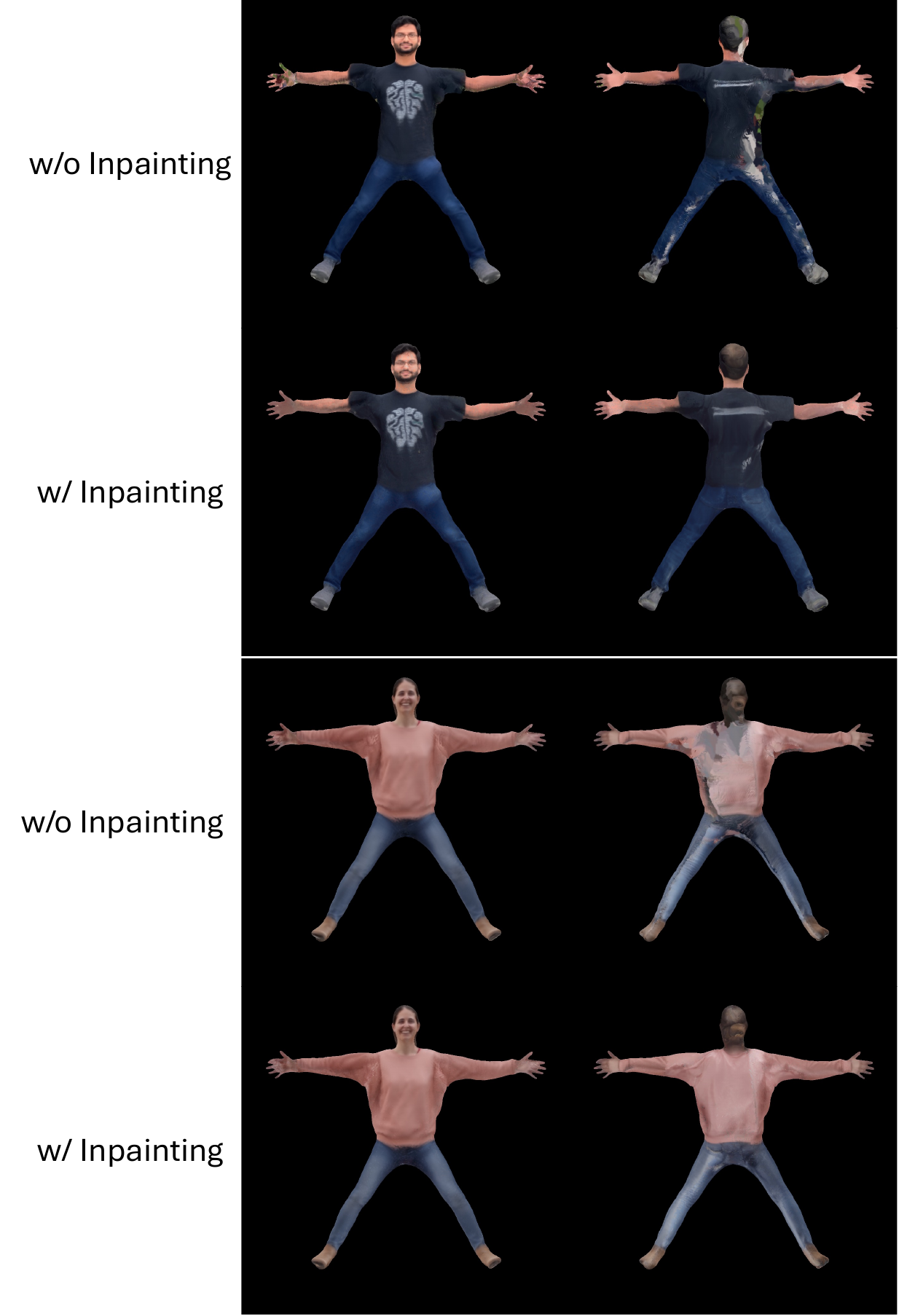}
% \fbox{\rule{0pt}{2in} \rule{.9\linewidth}{0pt}}

\caption{\textbf{Canonical texture inpainting.} Our diffusion-based inpainting module can effectively complete the textures that are missing from the monocular observations.}
\label{fig:cano_tex}
\end{figure}
}

\newcommand{\figureloose}{

\begin{figure}[t]
\raggedleft
\includegraphics[width=\linewidth,trim=0 0 0 0,clip]{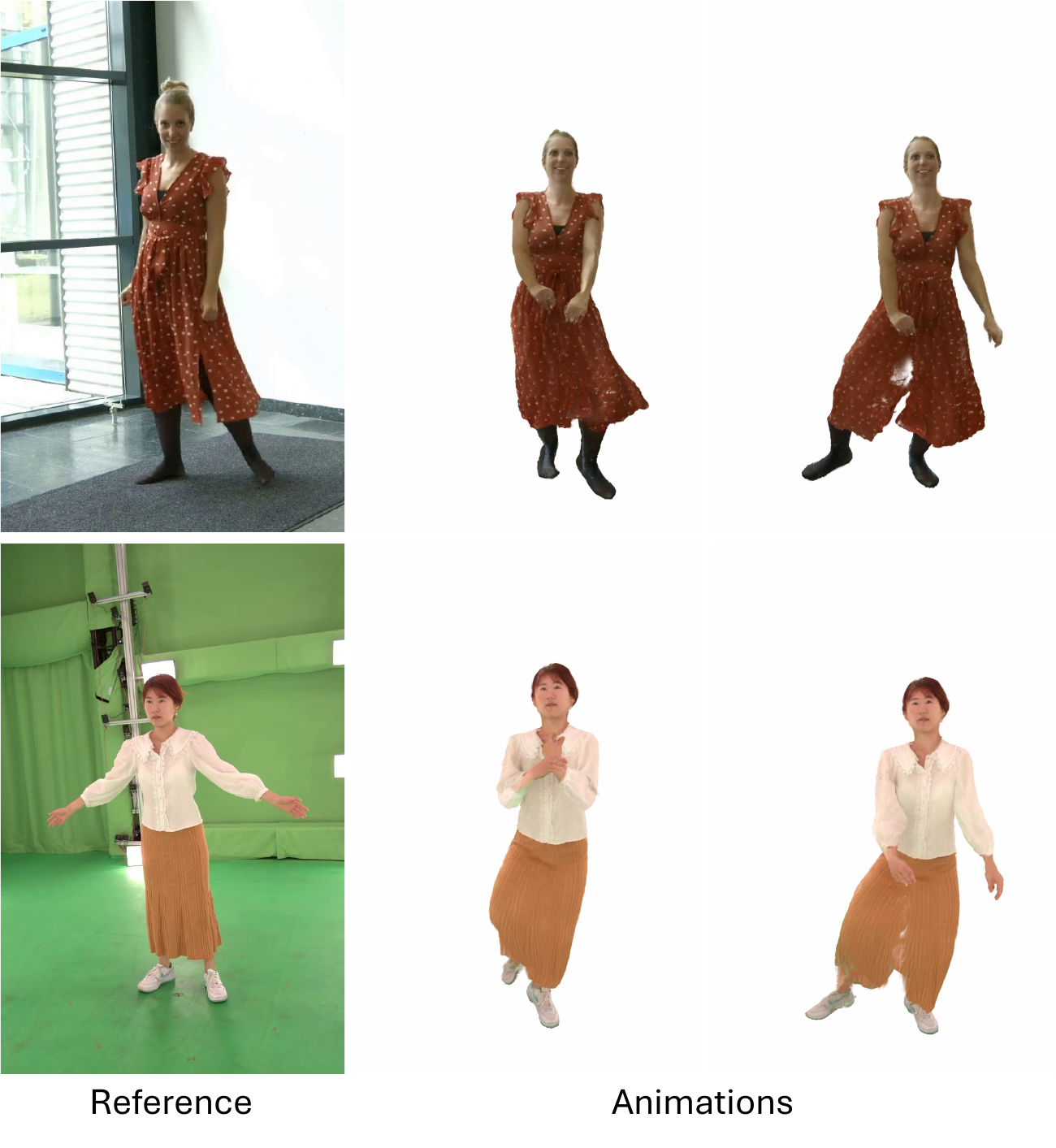}
% \fbox{\rule{0pt}{2in} \rule{.9\linewidth}{0pt}}

\caption{\textbf{Results on loose outfits.} Our method generates plausible renderings for less challenging driving signals but fail to output promising results for challenging human poses.}
\label{fig:loose}
\end{figure}
}

\newcommand{\figurelighting}{

\begin{figure}[t]
\raggedleft
\includegraphics[width=\linewidth,trim=0 0 0 0,clip]{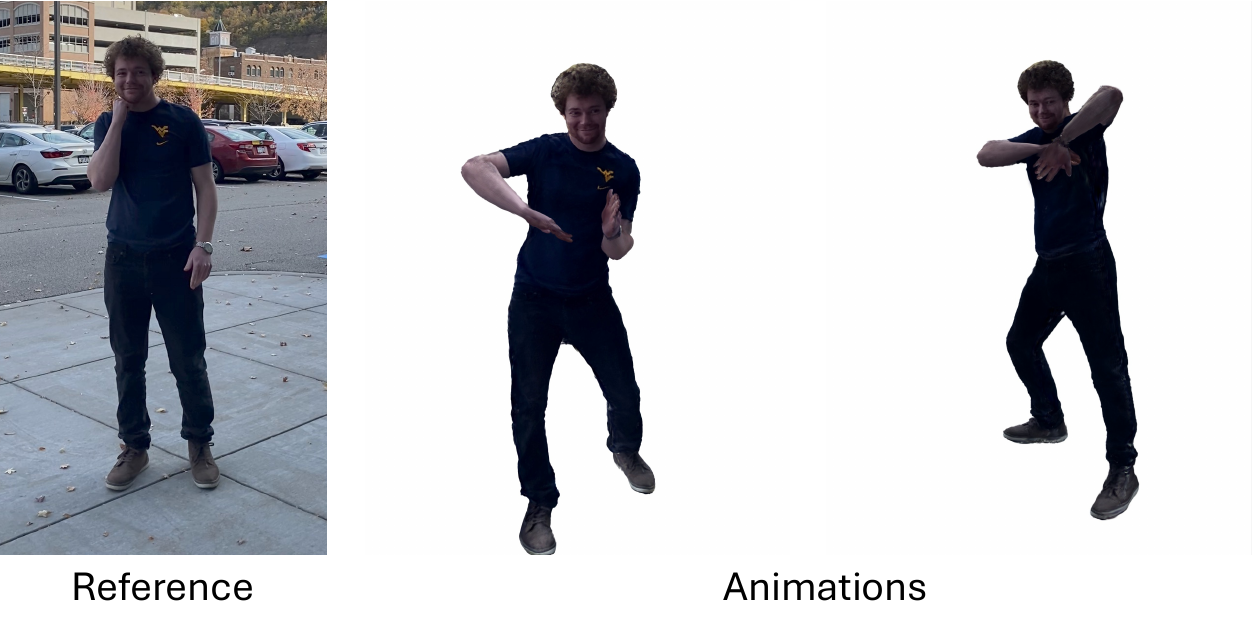}
% \fbox{\rule{0pt}{2in} \rule{.9\linewidth}{0pt}}

\caption{\textbf{Results on extreme lighting.} The brightness of the created human avatars is in its imperfection in case of a dark capture environment.}
\label{fig:lighting}
\end{figure}
}

\newcommand{\figuresuppexavatar}{

\begin{figure*}[t]
\raggedleft
\includegraphics[width=\linewidth,trim=0 0 0 0,clip]{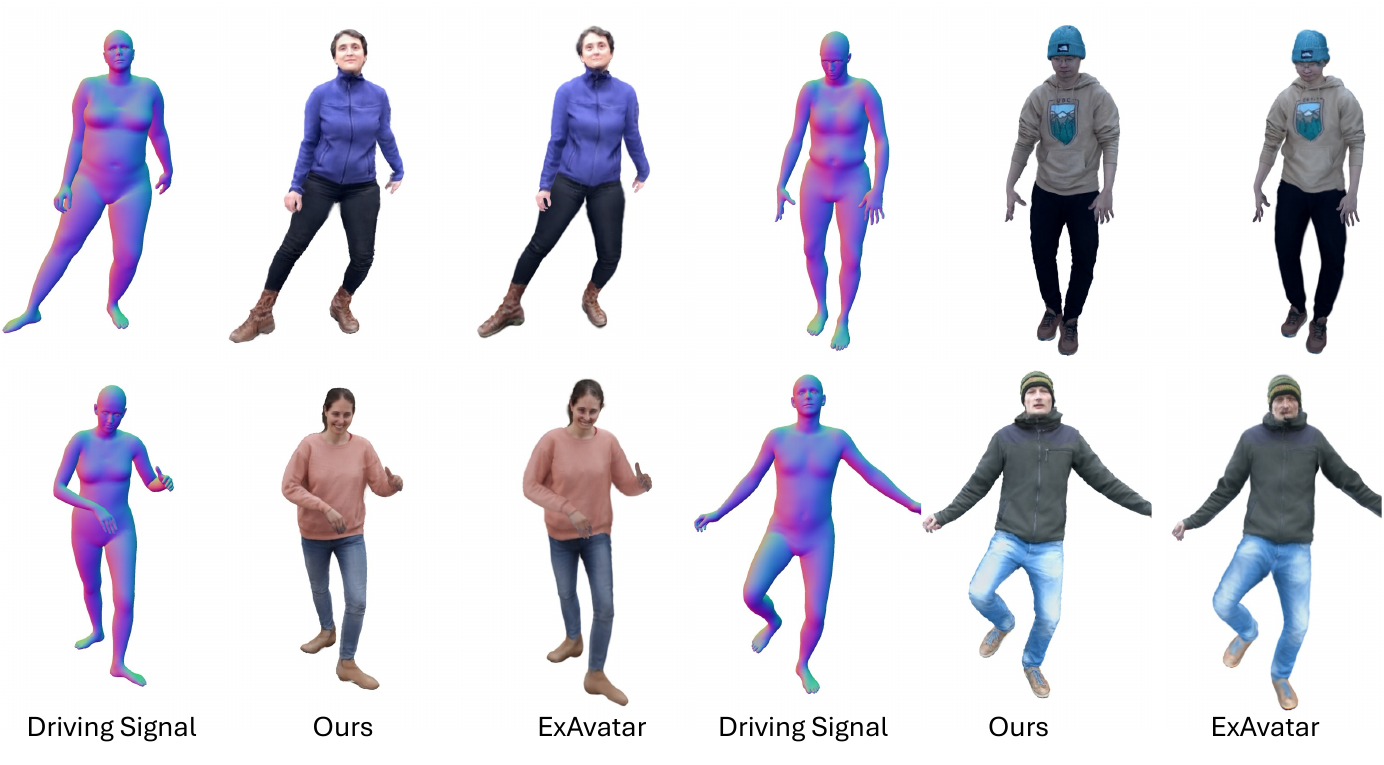}
% \fbox{\rule{0pt}{2in} \rule{.9\linewidth}{0pt}}

\caption{\textbf{Additional comparisons with ExAvatar.} Compared to ExAvatar, our method creates higher-quality 3D human avatars with finer-grained appearance details (\eg, clothing wrinkles and facial features), and generalizes better to out-of-distribution driving signals.}
\label{fig:suppexavatar}
\end{figure*}
}

\newcommand{\figureablationid}{

\begin{figure*}[t]
\raggedleft
\includegraphics[width=\linewidth,trim=0 0 0 0,clip]{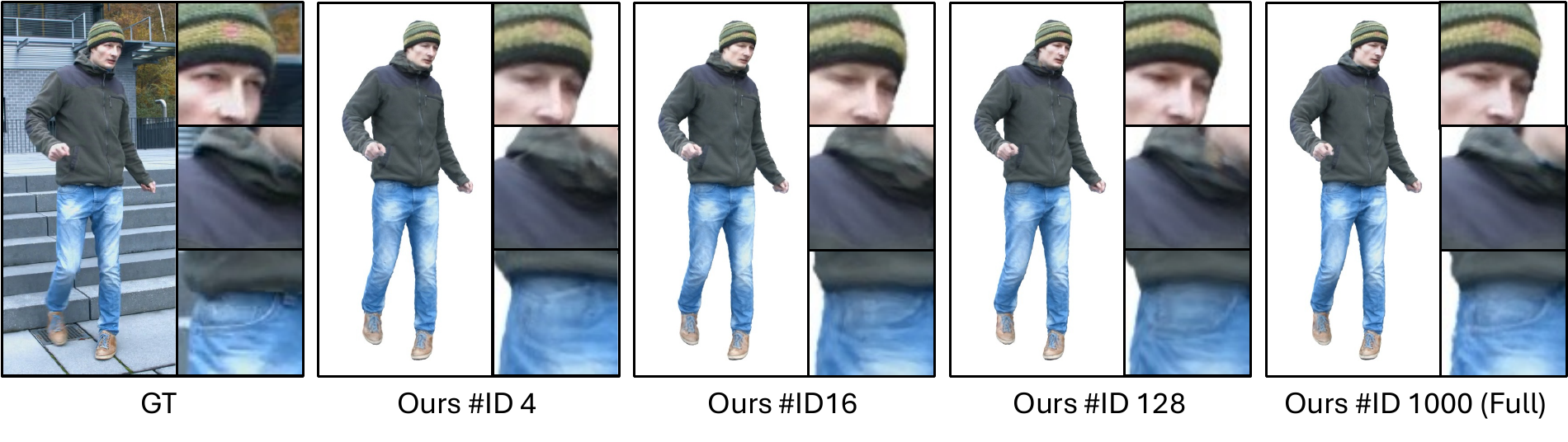}
% \fbox{\rule{0pt}{2in} \rule{.9\linewidth}{0pt}}

\caption{\textbf{Number of training IDs.} The rendering quality consistently increases when the universal prior model is trained on more identities/data. Especially, more appearance details can be recovered, \eg, the eyes and the pocket of the jeans, and our full model trained on 1000 subjects do not suffer from the problem caused by inaccurate opacity predictions, \cf the collar of the hoodie.}
\label{fig:ablationid}
\end{figure*}
}

\newcommand{\figuresuppquali}{

\begin{figure*}[t]
\raggedleft
\includegraphics[width=\linewidth,trim=0 0 0 0,clip]{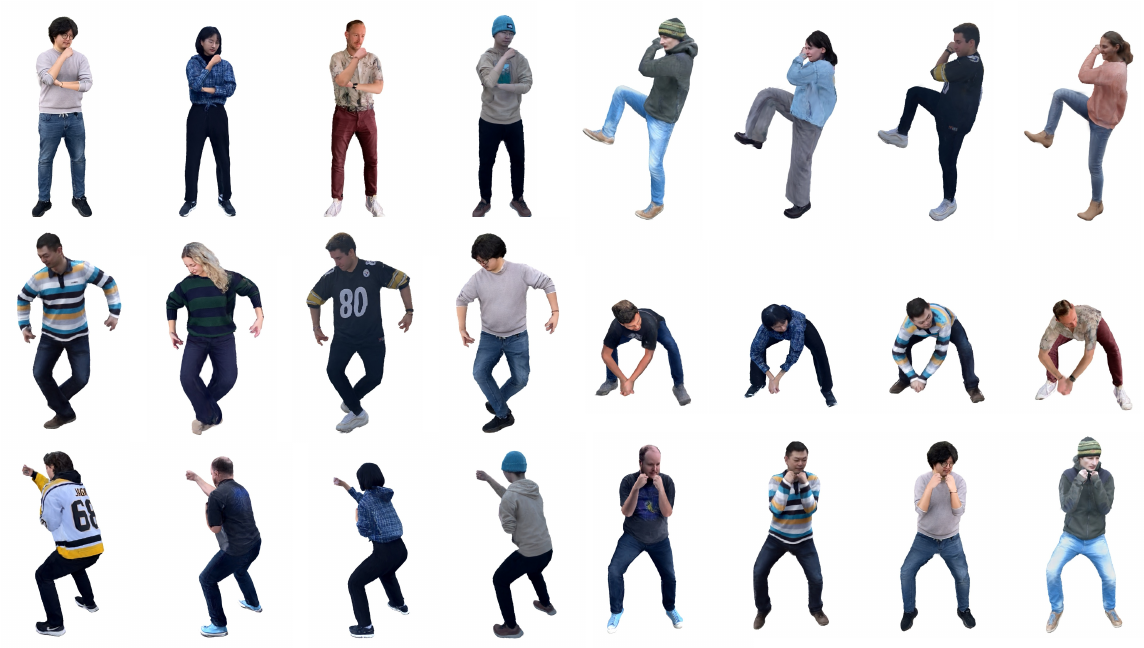}
% \fbox{\rule{0pt}{2in} \rule{.9\linewidth}{0pt}}

\caption{\textbf{Additional Visual animation results of avatars created from monocular in-the-wild videos.} The created 3D avatars can be animated using novel human poses and demonstrate highly detailed appearance from arbitrary view points.}
\label{fig:suppquali}
\end{figure*}
}

\newcommand{\tableneuman}{
\begin{table}[t]
% \small
\centering
\caption{\textbf{Quantitative interpolation synthesis comparisons on NeuMan dataset.} Our method consistently outperforms all baselines on all evaluation metrics (\cf \figref{fig:neuman}).}
\begin{tabular}{lccc}

\hline  Method & $\mathbf{PSNR} \uparrow$ & $\mathbf{SSIM} \uparrow$ & $\mathbf{LPIPS} \downarrow$ \\
\hline
HumanNeRF \cite{weng_humannerf_2022_cvpr} & 27.06 & 0.967 & 1.92 \\
InstantAvatar \cite{jiang2023instantavatar}  & 28.47 & 0.972 & 2.77 \\
NeuMan \cite{jiang2022neuman} & 25.48 & 0.966 & 2.87 \\
Vid2Avatar \cite{guo2023vid2avatar}  & 26.87 & 0.969 & 2.41 \\
3DGS-Avatar \cite{qian20233dgsavatar}  & 29.75 & 0.975 & 1.75 \\
GaussianAvatar \cite{hu2024gaussianavatar}  & \cellcolor{yyellow}29.94 & \cellcolor{yyellow}0.980 & \cellcolor{oorange}1.24 \\
ExAvatar \cite{moon2024exavatar} & \cellcolor{oorange}31.39 & \cellcolor{oorange}0.981 & \cellcolor{yyellow}1.64 \\
\hline
  Ours & \cellcolor{rred}32.71 & \cellcolor{rred}0.983 & \cellcolor{rred}1.19 \\
\hline

\end{tabular}
\vspace{-0.2cm}
\label{tab:neuman}
\end{table}
}

\newcommand{\tablemonoperfcap}{
\begin{table}[t]
% \small
\centering
\caption{\textbf{Quantitative extrapolation synthesis comparisons on MonoPerfCap dataset.} Our method consistently outperforms all baselines on all evaluation metrics (\cf \figref{fig:monoperfcap}).}
\begin{tabular}{lccc}

\hline  Method & $\mathbf{PSNR} \uparrow$ & $\mathbf{SSIM} \uparrow$ & $\mathbf{LPIPS} \downarrow$ \\
\hline
Vid2Avatar \cite{guo2023vid2avatar}  & 28.49 & 0.976 & 2.46 \\
ExAvatar \cite{moon2024exavatar} & \cellcolor{yyellow}30.29 & \cellcolor{oorange}0.979 & \cellcolor{yyellow}2.19 \\

\hline
  Ours w/o Fine-Tuning & 29.24 & \cellcolor{yyellow}0.977 & 2.29 \\
  Ours w/o Avg. Scale  & \cellcolor{oorange}31.03 & \cellcolor{oorange}0.979 & \cellcolor{oorange}1.51 \\
  Ours & \cellcolor{rred}31.97 & \cellcolor{rred}0.981 & \cellcolor{rred}1.37 \\
\hline

\end{tabular}
\vspace{-0.2cm}
\label{tab:monoperfcap}
\end{table}
}

\newcommand{\tableablateid}{
\begin{table}[t]
% \small
\centering
\caption{\textbf{Number of training IDs.} Rendering quality consistently increases when the prior model is trained on more identities.}
\begin{tabular}{lccc}

\hline  Method & $\mathbf{PSNR} \uparrow$ & $\mathbf{SSIM} \uparrow$ & $\mathbf{LPIPS} \downarrow$ \\
\hline
  Ours \#ID 4 & \cellcolor{yyellow}31.28 & \cellcolor{yyellow}0.979 & 1.53 \\
  Ours \#ID 16 & \cellcolor{yyellow}31.28 & \cellcolor{oorange}0.980 & \cellcolor{yyellow}1.51 \\
  Ours \#ID 128 & \cellcolor{oorange}31.34 & \cellcolor{oorange}0.980 & \cellcolor{oorange}1.45 \\
  Ours \#ID 1000 (Full) & \cellcolor{rred}31.97 & \cellcolor{rred}0.981 & \cellcolor{rred}1.37 \\
\hline

\end{tabular}
\vspace{-0.2cm}
\label{tab:ablationid}
\end{table}
}

\newcommand{\tableinpainting}{
\begin{table}[t]
% \small
\centering
\caption{\textbf{Importance of inpainting.} Our diffusion-based inpainting module can effectively complete the textures that are missing from the monocular observations (\cf. \figref{fig:cano_tex}).}
\begin{tabular}{lccc}

\hline  Method & $\mathbf{PSNR} \uparrow$ & $\mathbf{SSIM} \uparrow$ & $\mathbf{LPIPS} \downarrow$ \\
\hline

  Ours w/o Inpainting & 30.17 & 0.977 & 2.22 \\
  Ours & 30.22 & 0.977 & 2.18 \\
\hline

\end{tabular}
\vspace{-0.2cm}
\label{tab:inpainting}
\end{table}
}

\newcommand{\tablerebuttalmono}{
\begin{table}[t]
% \small
\centering
\caption{\textbf{Additional quantitative extrapolation synthesis comparisons on MonoPerfCap dataset.}}
\begin{tabular}{lccc}

\hline  Method & $\mathbf{PSNR} \uparrow$ & $\mathbf{SSIM} \uparrow$ & $\mathbf{LPIPS} \downarrow$ \\
\hline
InstantAvatar [24] & 25.92 & 0.968 & 3.73 \\
Vid2Avatar [13]  & 28.49 & 0.976 & 2.46 \\
3DGS-Avatar [56] & \cellcolor{yyellow}29.27 & \cellcolor{yyellow}0.977 & \cellcolor{oorange}1.80 \\
GaussianAvatar [18] & 27.22 & 0.974 & 2.27 \\
ExAvatar [47] & \cellcolor{oorange}30.29 & \cellcolor{oorange}0.979 & \cellcolor{yyellow}2.19 \\

\hline
  Ours & \cellcolor{rred}31.97 & \cellcolor{rred}0.981 & \cellcolor{rred}1.37 \\
\hline

\end{tabular}
\vspace{-0.3cm}
\label{tab:rebuttalmono}
\end{table}
}
\begin{abstract}
\vspace{-2em}

We present \methodname, a method to create photorealistic and animatable 3D human avatars from monocular in-the-wild videos. Building a high-quality avatar that supports animation with diverse poses from a monocular video is challenging because the observation of pose diversity and view points is inherently limited. The lack of pose variations typically leads to poor generalization to novel poses, and avatars can easily overfit to limited input view points, producing artifacts and distortions from other views. In this work, we address these limitations by leveraging a universal prior model (UPM) learned from a large corpus of multi-view clothed human performance capture data. We build our representation on top of expressive 3D Gaussians with canonical front and back maps shared across identities. Once the UPM is learned to accurately reproduce the large-scale multi-view human images, we fine-tune the model with an in-the-wild video via inverse rendering to obtain a personalized photorealistic human avatar that can be faithfully animated to novel human motions and rendered from novel views. The experiments show that our approach based on the learned universal prior sets a new state-of-the-art in monocular avatar reconstruction by substantially outperforming existing approaches relying only on heuristic regularization or a shape prior of minimally clothed bodies (\eg, SMPL) on publicly available datasets.

\end{abstract}    
\section{Introduction}

Authentic digital humans are widely used for the synthesis of novel animations of real persons in games and movies. They are also expected to be an indispensable component for virtual communications with immersive display devices (AR/VR).
However, creating such an authentic avatar typically requires expensive multi-view capture systems~\cite{bagautdinov2021driving}, limiting its availability to professional studios. If we can effortlessly build a high-quality avatar just from an in-the-wild monocular video, it unlocks myriads of applications not only for professionals but also for everybody.

Despite its promise, it remains non-trivial to create a high-quality avatar from in-the-wild videos that can be faithfully animated with diverse poses and efficiently rendered from arbitrary views. Monocular videos often do not cover the entire space of pose and view points required for test-time animation and rendering. The scarcity of pose variations leads to poor generalization to novel poses. Specifically, the avatar may exhibit artifacts and unnatural deformations when animated using out-of-distribution poses. Moreover, inverse rendering with limited view coverage is prone to overfitting, resulting in distortions and artifacts when rendered from unseen camera views.

Existing approaches~\cite{guo2023vid2avatar, jiang2022neuman, yu2023monohuman, weng_humannerf_2022_cvpr, moon2024exavatar, qian20233dgsavatar, Hu_2024_CVPR} aim to overcome its limitations by incorporating statistical prior from a minimally-clothed human body model~\cite{loper2015smpl, SMPL-X:2019} or geometric prior based on heuristics (\eg, Laplacian regularization~\cite{moon2024exavatar}). While these priors improve robustness under this ill-posed problem setting, there remains a clear quality gap from 3D human avatars reconstructed from high-quality studio data especially when animated with novel poses.

In this work, we argue that the core problem lies in the fact that the aforementioned priors are not built for clothed human avatar modeling. Parametric models such as SMPL~\cite{loper2015smpl} focus on minimally clothed bodies without appearance. Laplacian regularization uniformly penalizes deformations regardless of the underlying materials. To address this, we introduce \methodname, a method to create photorealistic 3D human avatars from monocular in-the-wild videos by leveraging universal prior directly learned from high-quality clothed human performance capture data.

Our universal prior model is based on 3D Gaussians~\cite{kerbl3Dgaussians} due to its efficiency and expressiveness in representing details. While existing universal models for face and hands~\cite{10.1145/3528223.3530143, li2024uravatar, chen2024urhand} rely on shared UV parameterization, it is not suitable for clothed humans due to the diverse topologies of clothing. Inspired by~\cite{li2024animatablegaussians}, we propose, for the first time, front and back maps as a universal parametrization for clothed humans. More specifically, we warp static SDF-based reconstructions~\cite{yariv2021volume, guo2023vid2avatar} of all subjects to a canonical space by normalizing both poses and bone lengths, and obtain color and position maps from front and back views with orthogonal projection. This allows us to maximize the spatial alignment of identity conditioning data across identities while supporting diverse topologies of clothing. We feed the front and back maps to a U-Net~\cite{10.1145/3528223.3530143, li2024uravatar, unet} to predict pose-dependent Gaussian attributes in a pixel-aligned manner with the input maps. We train this universal prior model via inverse rendering with multi-view performance capture of a thousand of clothed humans.
Once trained, we personalize the universal prior model to monocular in-the-wild videos.
We first reconstruct the canonical textured template from the monocular observations using neural SDF~\cite{yariv2021volume, guo2023vid2avatar}.
Given the conditioning data obtained from the canonical textured template, we inpaint the unseen texture regions using a diffusion model~\cite{esser2024scaling} adapted for inpainting. We then fine-tune the model via inverse rendering and update the network weights of the prior model to faithfully reconstruct person-specific details from the monocular observations.

The experiments show that our approach produces photorealistic avatars that can be animated beyond training pose distributions with faithful pose-dependent deformations and appearance changes (\cf \figref{fig:teaser}). We also carefully validate our design choices via ablation studies. Furthermore, we compare our method with state-of-the-art approaches in novel view/pose synthesis using publicly available datasets, and demonstrate that \methodname outperforms them by a substantial margin both quantitatively and qualitatively. 

In summary, our contributions are:
\begin{compactitem}
 \item A universal prior model of clothed humans directly learned from a thousand of high-quality dynamic performance capture data.
 \item The first universal prior model architecture designed to use spatially normalized front and back identity conditioning data, enabling efficient and scalable training of multi-identity clothed human avatars.
 \item A robust personalization pipeline to create photorealistic and animatable clothed human avatars from monocular in-the-wild videos, achieving new state-of-the-art performance.
\end{compactitem}

\section{Related Work}

\paragraph{Avatar Reconstruction from Multi-View Videos.} High-fidelity 3D avatar reconstruction has required calibrated multi-view systems~\cite{liu2021neural, bagautdinov2021driving, remelli2022drivable, peng2021neural, habermann2021, 10.1145/3478513.3480545, peng2021animatable, xu2022sanerf, HVTR:3DV2022, 2021narf, li2022tava, zhang2021stnerf, ARAH:ECCV:2022, chen2024meshavatar, saito2024rgca, zheng2023avatarrex, li2023posevocab, 10.1145/3697140, shen2023xavatar, yin2023hi4d}. These approaches utilize neural rendering techniques (\eg,~\cite{mildenhall2020nerf, yariv2021volume}) to learn an implicit canonical representation of clothed humans.
More recently, researchers retain the human skeletal structure and substituting implicit neural rendering with 3DGS~\cite{kerbl3Dgaussians} to learn person-specific animatable avatars~\cite{li2024animatablegaussians, zielonka2023drivable3dgaussianavatars, moreau2024human, Pang_2024_CVPR, PhysAavatar24, jung2023deformable3dgaussiansplatting}. 
Our front and back map representation is inspired by~\cite{Pang_2024_CVPR, hu2024gaussianavatar, li2024animatablegaussians}, specifically Li~\etal~\cite{li2024animatablegaussians}. Unlike UV parameterization which necessitates manual efforts \cite{Pang_2024_CVPR} or only covers minimally-clothed bodies \cite{hu2024gaussianavatar}, our front and back parameterization can be obtained automatically and accommodates various topologies of clothed humans. Different from Li~\etal~\cite{li2024animatablegaussians}, who use front and back position maps as pose features for single identity modeling, we propose, for the first time, to extend such parameterization to be universal across multiple identities.
Compared to aforementioned methods that require a specialized capturing setup, our approach creates photorealistic and animatable avatars from just monocular RGB videos via learned universal prior.

\paragraph{Avatar Reconstruction from Monocular Video.} Traditional mesh-based methods~\cite{loper2015smpl, SMPL-X:2019, Xu:2018:MHP:3191713.3181973, deepcap, guo2021human, rong2021frankmocap} are limited to a fixed topology and resolution, and cannot represent fine-grained details like the human faces.
Regression-based methods that learn to regress 3D human shape and texture from images have demonstrated compelling results~\cite{xiu2022icon, xiu2023econ, saito2020pifuhd, huang2024tech, zheng2021pamir, saito2019pifu, He_2021_ICCV, alldieck2022phorhum, huang2020arch, ho2024sith, feng2023foflearningfourieroccupancy, xiu2024puzzleavatar}. 
A major limitation of these methods is that the reconstructed avatars can only be driven using rigid skeleton movement without pose-dependent deformations, leading to unrealistic animation results. 
Fitting articulated implicit neural fields to monocular videos via inverse neural rendering has been demonstrated~\cite{su2021anerf, su2022danbo, jiang2022selfrecon, yu2023monohuman, weng_humannerf_2022_cvpr, jiang2022neuman, guo2023vid2avatar, reloo, multiply, xue2024hsr, jiang2023instantavatar}.
Following this line of work, emerging methods fit articulated 3D Gaussians to the monocular videos by either directly optimizing the 3D Gaussian attributes~\cite{Lei_2024_CVPR, shao2024splattingavatar, svitov2024hahahighlyarticulatedgaussian, li2024gaussianbodyclothedhumanreconstruction} or training neural networks to predict the attributes~\cite{qian20233dgsavatar, wen2024gomavatar, kocabas2024hugs, Hu_2024_CVPR, hu2024gaussianavatar, li2023human101, moon2024exavatar, liu2024gvareconstructingvivid3d}. GaussianAvatar~\cite{hu2024gaussianavatar} leverages a UV positional map of SMPL to generate pose-dependent effects. ExAvatar~\cite{moon2024exavatar} combines with SMPL-X to achieve a whole-body avatar with face and hand control. These methods achieve improved rendering speed and quality. However, the reconstructed avatars still show blurriness and lack appearance details. Especially, they tend to generate uncanny animation results for poses that are out of the training distribution from the short video. In contrast, our method leverages a high-quality and large-scale corpus of dynamic human captures to train a universal prior model, which can be personalized to short video data and ensure superior generalization ability to novel poses.
\vspace{-1em}
\paragraph{Human-Centric Prior Model.}
Statistical body models~\cite{loper2015smpl, SMPL-X:2019, Joo_2018_CVPR} can hardly model human surface details such as clothing and facial features. Regression-based methods~\cite{saito2019pifu, saito2020pifuhd, xiu2022icon, xiu2023econ, ho2024sith, huang2020arch, He_2021_ICCV, chatziagapi2024migs} train networks to learn pixel-aligned human features. These models are trained on small-scale static 3D human scans, thus the reconstructed humans are often not animatable or can only show rigid and unrealistic animations. Generalizable human rendering methods are trained on small-scale multi-view images to synthesize human novel views from sparse camera inputs without supporting human controllability~\cite{kwon2021neural, kwon2024ghg, sun2024metacap, Zhao_2022_CVPR, Chen_2023_CVPR, chen2022gpnerf, zheng2024gpsgaussian}. There are some promising prior models that recover detailed appearance and generate plausible animations for human hands~\cite{Corona_2022_CVPR, Moon_2024_CVPR, chen2024urhand} and faces~\cite{10.1145/3528223.3530143, buhler2023preface, buehler2024cafca, li2024uravatar}. In this work, we propose a universal prior model for photorealistic and animatable clothed humans, learned from thousands of dynamic human performance captures with diverse identities, garment styles, view points and poses.
\figurePipeline
\section{Method}
We introduce \methodname, a method to create photorealistic 3D human avatars from monocular in-the-wild videos. 
Our method is illustrated in \figref{fig:pipeline} and consists of two main steps: building universal prior model for clothed humans (\secref{sec:upm}) and in-the-wild personalization (\secref{sec:personalization}).

\subsection{Universal Prior Model for Clothed Humans}
\label{sec:upm}
\paragraph{Geometry/Texture Representations.} We represent the geometry and texture of clothed humans as a set of 3D Gaussians defined in a canonical space shared across identities. Each Gaussian is denoted as:
\begin{equation}
\mathbf{g}=\{\mathbf{x}, \mathbf{q}, \mathbf{s}, o, \mathbf{c}\}. 
\end{equation}
The parameters include a position $\mathbf{x} \in \mathbb{R}^{3}$, a unit quaternion $\mathbf{q} \in \mathbb{R}^{4}$, a scale factor $\mathbf{s} \in \mathbb{R}^{3}_{+}$ along three orthogonal axis, an opacity value $o \in \mathbb{R}_{+}$ and a color $\mathbf{c} \in \mathbb{R}^{3}_{+}$. The associated covariance matrix can be calculated as $\mathbf{\Sigma}=\mathbf{R}\diag(\mathbf{s})\diag(\mathbf{s})^\top\mathbf{R}^\top$, where the rotation matrix $\mathbf{R}$ can be easily converted from $\mathbf{q}$.

We deform the canonical 3D Gaussians to the posed space via forward linear blend skinning (LBS). Given the human pose $\boldsymbol{\theta}$ and shape $\boldsymbol{\beta}$ (consolidated as $\boldsymbol{\Theta}$), defined analogously to SMPL-X~\cite{SMPL-X:2019}, we compute the forward transformation matrix $\mathbf{T}$ as follows:
\begin{equation}
\label{eq:lbs}
    \mathbf{T}=\sum_{i = 1}^{n_b} w_{c}^i \mathbf{B}_i,
\end{equation}
where $n_b$ denotes the number of bones in the transformation and $\mathbf{B}_i$ are the bone transformation matrices for joints $i \in \{1,...,n_{b}\}$, derived from the parameters $\boldsymbol{\Theta}$. $\mathbf{w}_{c}=\{w_{c}^1,...,w_{c}^{n_b}\}$ represents the skinning weights which are queried from a diffused skinning map $\mathcal{S}_c$ based on the canonical SMPL-X vertices~\cite{lin2022fite, SMPL-X:2019}. The position $\mathbf{x}_c$ and covariance $\mathbf{\Sigma}_c$ of canonical Gaussians can then be transformed to the deformed space accordingly as follows:
\begin{equation}
\label{eq:forward_lbs}
    \mathbf{x}_d = \mathbf{T}\mathbf{x}_c, \
    \mathbf{\Sigma}_d = \mathbf{T}_{1:3}\mathbf{\Sigma}_c\mathbf{T}^\top_{1:3}.
\end{equation}

\paragraph{Universal Identity Conditioning.}
We propose front and back maps as a universal parameterization for clothed humans. Such parameterization is obtained by orthogonally projecting the normalized canonical textured template into front and back views. Thus, we can obtain the front and back position maps $\mathcal{P}^\text{f}_c$ and $\mathcal{P}^\text{b}_c$, and texture maps $\mathcal{T}^\text{f}_c$ and $\mathcal{T}^\text{b}_c$ in canonical space. $\mathcal{P}^\text{f}_c$ and $\mathcal{P}^\text{b}_c$ are stored with the 3D positions of the canonical template. We abbreviate the concatenation of the canonical front and back position and texture maps as $\mathcal{P}_c$ and $\mathcal{T}_c$. 
For the sake of simplicity and clarity, we also adopt such abbreviation for other front and back maps.
We use $\mathcal{T}_c$ as the identity conditioning data, enabling the feasibility of training a universal prior model $\mathcal{F}$ in a cross-identity setting.
To accompany this, we define the output of our universal prior model $\mathcal{F}$ as a Gaussian map (front/back) $\mathcal{G}(\boldsymbol{\Theta})$, in which every pixel within the projected template mask $\mathcal{M}_c$ represents a 3D Gaussian with its attributes $\mathbf{g}$.

To obtain our conditioning data, we reconstruct the canonical template with vertices $\mathbb{V}$ via SDF-based volume rendering and LBS-based inverse mapping~\cite{yariv2021volume, guo2023vid2avatar} from multi-view images. The reconstruction is based on a single automatically selected keyframe in which the human pose $\boldsymbol{\theta}$ has the highest similarity to our pre-defined canonical human pose $\boldsymbol{\theta}_{\text{cano}}$. The texture is then obtained by unwrapping the RGB colors from the multi-view images onto the canonical template mesh. 

To facilitate better learning of pose-dependent surface deformations and appearance changes across large-scale human identities, irrespective of bone lengths or human scales, we normalize the canonical templates based on an average human skeleton scale. Specifically, we set the shape parameters $\boldsymbol{\beta} = \mathbf{0}$ and pose parameters $\boldsymbol{\theta} = \boldsymbol{\theta}_{\text{cano}}$ for all subjects and compute the normalization transformation matrix $\mathbf{T}_{\text{norm}}$ similar to \equref{eq:lbs}. We then apply the normalization to the canonical templates as follows:
\begin{equation}
\label{eq:normalization}
    \mathbf{v}_{\text{norm}} = \mathbf{T}_{\text{norm}}\mathbf{v}, \forall \ \mathbf{v} \in \mathbb{V}.
\end{equation}
Given the normalized canonical textured template, we generate spatially aligned identity conditioning data for all training subjects by orthogonally projecting the templates into front and back views, yielding the canonical template masks $\mathcal{M}_c$, position maps $\mathcal{P}_c$, and texture maps $\mathcal{T}_c$.

\paragraph{Universal Prior Model Training.} Inspired by prior work~\cite{10.1145/3528223.3530143, li2024uravatar}, we employ a U-Net as the universal model backbone~\cite{unet}. In particular, our universal model $\mathcal{F}$ takes the person-specific canonical texture maps $\mathcal{T}_c$ as identity conditioning input.
To model pose-dependent effects such as dynamically changing wrinkles on clothes, we further concatenate the posed position maps $\mathcal{P}_d(\boldsymbol{\Theta})$ as additional input for each training iteration to predict pose-dependent Gaussian maps $\mathcal{G}(\boldsymbol{\Theta})$. Similar to deforming the canonical 3D Gaussians to the posed space (\cf \equref{eq:forward_lbs}), we obtain the posed position maps $\mathcal{P}_d(\boldsymbol{\Theta})$ by applying forward LBS to the 3D positions stored in each valid pixel of the canonical position maps $\mathcal{P}_c$, excluding the global orientation and translation which do not affect the human dynamic details. Thus, our universal model works as follows:
\begin{equation}
    \mathcal{F}(\mathcal{T}_c, \mathcal{P}_d(\boldsymbol{\Theta})) \rightarrow \mathcal{G}(\boldsymbol{\Theta}).
\end{equation}

We extract the Gaussian attributes $\mathbf{g}$ for all valid pixels within $\mathcal{M}_c$ from the pose-dependent Gaussian maps $\mathcal{G}(\boldsymbol{\Theta})$. In practice, we opt to predict the position and color offsets $\Delta\mathbf{x}(\boldsymbol{\Theta}), \Delta\mathbf{c}(\boldsymbol{\Theta})$ relative to the canonical maps instead of absolute position/color maps. This encourages our model to focus on learning finer-grained details within a given model capacity. Thus, we attain canonical 3D Gaussians with positions $\mathbf{x}=\hat{\mathbf{x}}+\Delta\mathbf{x}(\boldsymbol{\Theta})$ and colors $\mathbf{c}=\hat{\mathbf{c}}+\Delta\mathbf{c}(\boldsymbol{\Theta})$, where $\hat{\mathbf{x}}$ and $\hat{\mathbf{c}}$ are queried from $\mathcal{P}_c$ and $\mathcal{T}_c$.

After applying \equref{eq:forward_lbs}, we obtain all 3D Gaussians in the posed space. Following splatting-based rasterization~\cite{kerbl3Dgaussians}, the pixel color $\mathbf{C}$ is obtained by the $\alpha$-blending of $N$ overlapping Gaussians that are depth-ordered in posed space:
\begin{equation}
\mathbf{C}=\sum_{i=1}^N \mathbf{c}^i \alpha^i \prod_{j=1}^{i-1}\left(1-\alpha^j\right),
\end{equation}
where the transparency $\alpha$ is evaluated using the 2D covariance projected from $\mathbf{\Sigma}$ and multiplied with the per-Gaussian opacity $o$. Please refer to~\cite{kerbl3Dgaussians} for more details.

\paragraph{Training Objectives.} The training objectives for our universal prior model include L1 and perceptual losses~\cite{zhang2018perceptual} between the rendered and ground-truth images, along with an offset regularization loss:
\begin{equation}
\label{eq:loss}
\mathcal{L} = 
\mathcal{L}_{\text{1}} + 
\lambda_{\text{lpips}} \mathcal{L}_{\text{lpips}} +
\lambda_{\text{offset}} \mathcal{L}_{\text{offset}},
\end{equation}
where $\lambda_{\{\cdot\}}$ denote loss weights, and the regularization loss $\mathcal{L}_{\text{offset}}$ penalizes excessively large per-Gaussian offsets $\Delta\mathbf{x}(\boldsymbol{\Theta})$. We minimize the loss function $\mathcal{L}$ across all multi-view clothed human performance capture data.

\subsection{In-the-Wild Personalization}
\label{sec:personalization}
\paragraph{Preprocessing.} 
Given the monocular in-the-wild video, we first track the human shape and poses $\boldsymbol{\Theta}$ using an off-the-shelf SMPL-X estimator~\cite{Sun_2024_CVPR}.
We then employ Sapiens~\cite{10.1007/978-3-031-73235-5_12} to predict the 2D keypoints which are used to further refine the initial pose/shape estimates by minimizing the 2D keypoint projection error.
These 2D keypoints are served as point prompts for SAM~\cite{Kirillov_2023_ICCV, sam_hq} to acquire the human masks.
Given the pose/shape estimation and segmentation masks, we employ a method akin to that used for studio data to reconstruct the 3D canonical textured template.
Similarly, we follow the skeleton-based normalization strategy to attain the spatially aligned identity conditioning data. 
These steps serve to mitigate the domain gap in conditioning data between in-the-wild video sequences and the multi-view training data of our universal prior model. 
More details can be found in the \suppmat.

\paragraph{Diffusion-based Texture Inpainting.}
Unlike the multi-view capture data, human performance in in-the-wild videos exhibits varying levels of visibility, rather than consistent full-body visibility.
Hence, we develop a latent diffusion model to inpaint the textures of canonical templates.

Instead of inpainting the textured template in 3D, we formulate this problem as a 2D inpainting task applied to the unwrapped canonical texture maps. 
We begin by generating various visibility masks through rasterization of the pre-acquired studio canonical templates using randomly positioned sparse cameras.
We then fine-tune a pre-trained latent diffusion model with a DiT-based~\cite{peebles2023scalable} architecture similar to that of~\cite{esser2024scaling} using the generated visibility masks and canonical texture maps derived from the studio data.
Specifically, we train our diffusion model to denoise the canonical texture maps given partially masked inputs and their corresponding visibility masks.
During inference, we obtain the visibility mask by rasterizing the template using the monocular input views and use it to generate a complete canonical texture map. 
Note that our diffusion model jointly operates on the front and back texture maps and can inpaint missing regions with plausible textures by utilizing context from both sides.
We refer to the \suppmat for more details.

\paragraph{Fine-Tuning.}
To create 3D avatars with identity-preserving details while maintaining the generalization power to novel human poses, we further fine-tune our universal prior model on the monocular video observations via inverse rendering with a small number of iterations.
Due to the potentially inaccurate pose initialization, we fine-tune all network parameters, and optimize the estimated human shape and per-frame pose parameters $\boldsymbol{\Theta}$ jointly.
The loss function employed during the fine-tuning stage mirrors that utilized in the universal model training stage (\cf \equref{eq:loss}).
\section{Experiments}
We first introduce the training corpus and the test datasets. Then we compare \methodname with state-of-the-art approaches in two tasks: interpolation and extrapolation synthesis. Ablation studies are then conducted to show the effectiveness of our core components and design choices. Finally, we demonstrate more animation results of avatars created from in-the-wild monocular videos qualitatively.
\subsection{Training Dataset}
\label{sec:corpus}
We use a multi-view system similar to \cite{bagautdinov2021driving} to capture dynamic human performance, where we obtain calibrated and synchronized multi-view images at a resolution of 4096$\times$2668 pixels through the use of 200 cameras.
Participants are instructed to perform casual human motions for an average of approximately 5,000 frames per person. In total, 1,000 participants are recorded for model training.

\subsection{Test Datasets}
\paragraph{NeuMan Dataset \cite{jiang2022neuman}:} This dataset comprises a collection of videos captured by a mobile phone, featuring a single person performing a walking motion.
Following previous works \cite{hu2024gaussianavatar, moon2024exavatar}, we use \textit{bike}, \textit{citron}, \textit{jogging}, and \textit{seattle} sequences for comparisons that exhibit most body regions and contain minimally blurry images.
Since the official training and testing splits have interwoven frames,
we regard the evaluation on NeuMan dataset as an interpolation view synthesis task.

\paragraph{MonoPerfCap Dataset \cite{Xu:2018:MHP:3191713.3181973}:} This dataset contains in-the-wild videos of people with different garment types and various daily actions.
Compared to NeuMan dataset \cite{jiang2022neuman}, MonoPerfCap dataset includes more clothing deformations and pose diversity.
We use the first 80\% of each captured video for training and the remaining 20\% frames for testing. Thus, we define the evaluation on MonoPerfCap dataset as a more challenging extrapolation view synthesis task.

\tableneuman
\figureneuman
\paragraph{Evaluation Protocol:} We report PSNR, SSIM \cite{wang2003multiscale}, and LPIPS ($\times$100) \cite{zhang2018perceptual} for all synthesis comparisons. In all evaluation tables, the top three techniques are highlighted in \textcolor{rred}{red}, \textcolor{oorange}{orange}, and \textcolor{yyellow}{yellow}, respectively.

\subsection{Interpolation Synthesis Comparisons}
We conduct interpolation synthesis comparisons on NeuMan dataset \cite{jiang2022neuman}. For methods that jointly model the human and the background, we only compare the foreground (human) rendering quality. The quantitative results are partially sourced from \cite{hu2024gaussianavatar, moon2024exavatar}. Overall, our method outperforms all baseline methods in the interpolation setting by a substantial margin on all metrics (\cf \tabref{tab:neuman}). The difference is more visible in the qualitative comparison shown in \figref{fig:neuman}, where baseline methods cannot produce sharp details such as the wrinkles on the clothing and zippers, and tend to generate artifacts on the face or feet. In contrast, \methodname recovers more surface details (\eg, clothing wrinkles and facial features). We attribute this to our expressive universal prior model that is trained on high-quality data.

\tablemonoperfcap
\figuremonoperfcap
\subsection{Extrapolation Synthesis Comparisons}
We further conduct a more challenging experiment: extrapolation synthesis comparisons on MonoPerfCap dataset \cite{Xu:2018:MHP:3191713.3181973}.
Despite a more challenging setting, \tabref{tab:monoperfcap} indicates that our method still outperforms baselines across all metrics. This disparity becomes more evident in qualitative comparisons presented in \figref{fig:monoperfcap}. ExAvatar \cite{moon2024exavatar} struggles to recover the accurate facial features, and both ExAvatar \cite{moon2024exavatar} and Vid2Avatar \cite{guo2023vid2avatar} produce blurry renderings (\cf the shoes in the first row of \figref{fig:monoperfcap}). Thanks to our pre-trained universal prior model, \methodname can generalize better to out-of-distribution poses with plausible pose-dependent surface/appearance deformations (\cf the hem in the second row of \figref{fig:monoperfcap}). Our method also consistently preserves superior photorealism compared to the baselines.

\subsection{Ablation Studies}

\tableablateid
\paragraph{Training Data.} A key factor that affects the performance of our universal prior model is the number of training identities. We quantitatively analyze the effects by training the universal prior model with different numbers of training subjects, \ie $n_p \in \{4,16,128,1000\}$. The training subjects are chosen at random. We use MonoPerfCap dataset to evaluate the extrapolation synthesis performance. The quantitative results are reported in \tabref{tab:ablationid}. We observe that the final rendering quality consistently improves as the amount of training data/identities increases.

\figureavgscale
\paragraph{Conditioning Data Normalization.} To investigate the importance of skeleton-based conditioning data normalization, we compare our full model to a version without normalizing using an average scale (w/o Avg. Scale). We demonstrate in \tabref{tab:monoperfcap} that skeleton-based normalization helps to improve the generalization ability of our model to unseen identities and challenging poses (\cf \figref{fig:ablation_avg_scale}).

\figureinpainting
\paragraph{Diffusion-based Texture Inpainting.} We introduce a diffusion-based canonical inpainting module to complete the missing textures that are not visible in the monocular in-the-wild videos. As illustrated in \figref{fig:ablation_inpainting}, our inpainting module effectively eliminates artifacts caused by missing observations and inpaint the 3D avatar with plausible textures. More examples can be found in the \suppmat.

\figurequali
\figurefinetune
\paragraph{Fine-Tuning.} Due to the inherent domain gap between in-the-wild videos and studio data, fine-tuning the prior model based on monocular observations is crucial to recovering person-specific details. To demonstrate the effectiveness, we conduct an ablation experiment without fine-tuning on MonoPerfCap dataset (w/o Fine-Tuning). Results in \tabref{tab:monoperfcap} indicate that without fine-tuning, the final renderings are blurrier and lack detailed appearance features, as shown in \figref{fig:ablation_finetune} (\eg, the T-shirt pattern and the belt).

\subsection{Qualitative Results}
In \figref{fig:quali_res}, we showcase the animation results of avatars created from in-the-wild monocular videos. \methodname can generalize to different identities and garment styles, and produce highly realistic renderings for novel human poses and arbitrary view points. More qualitative results animated by various poses are available in the \suppmat.
\section{Conclusion}
In this work, we present \methodname to create photorealistic and animatable 3D avatars from monocular in-the-wild videos via universal prior. We first propose a universal parameterization for clothed humans that enables cross-identity training schema. We then build a photorealistic and animatable universal prior model for clothed humans learned from a thousand of high-quality dynamic performance capture data. Finally, we construct a robust pipeline that supports personalization of the universal prior model to monocular in-the-wild videos and create personalized photorealistic human avatars that can be faithfully animated to novel human motions and rendered from novel view points.

\noindent\textbf{Limitations and Future Works:} The current training dataset for our universal prior model lacks diverse facial expressions and dynamic capture data of human subjects dressed in loose garments. Therefore, our method currently does not support animatable faces or loose clothing. Additionally, our method assumes standard lighting conditions and may not perform optimally in environments with extreme lighting variations. We refer to \suppmat for a more detailed discussion of limitations and societal impact.

\clearpage
{
    \small
    \bibliographystyle{ieeenat_fullname}
    \bibliography{main}

\begin{thebibliography}{96}
\providecommand{\natexlab}[1]{#1}
\providecommand{\url}[1]{\texttt{#1}}
\expandafter\ifx\csname urlstyle\endcsname\relax
  \providecommand{\doi}[1]{doi: #1}\else
  \providecommand{\doi}{doi: \begingroup \urlstyle{rm}\Url}\fi

\bibitem[Alldieck et~al.(2022)Alldieck, Zanfir, and Sminchisescu]{alldieck2022phorhum}
Thiemo Alldieck, Mihai Zanfir, and Cristian Sminchisescu.
\newblock Photorealistic monocular 3d reconstruction of humans wearing clothing.
\newblock In \emph{Proceedings of the IEEE/CVF Conference on Computer Vision and Pattern Recognition (CVPR)}, 2022.

\bibitem[Bagautdinov et~al.(2021)Bagautdinov, Wu, Simon, Prada, Shiratori, Wei, Xu, Sheikh, and Saragih]{bagautdinov2021driving}
Timur Bagautdinov, Chenglei Wu, Tomas Simon, Fabi{\'a}n Prada, Takaaki Shiratori, Shih-En Wei, Weipeng Xu, Yaser Sheikh, and Jason Saragih.
\newblock Driving-signal aware full-body avatars.
\newblock \emph{ACM Transactions on Graphics (TOG)}, 40\penalty0 (4):\penalty0 1--17, 2021.

\bibitem[Buehler et~al.(2024)Buehler, Li, Wood, Helminger, Chen, Shah, Wang, Garbin, Orts-Escolano, Hilliges, Lagun, Riviere, Gotardo, Beeler, Meka, and Sarkar]{buehler2024cafca}
Marcel~C. Buehler, Gengyan Li, Erroll Wood, Leonhard Helminger, Xu Chen, Tanmay Shah, Daoye Wang, Stephan Garbin, Sergio Orts-Escolano, Otmar Hilliges, Dmitry Lagun, Jérémy Riviere, Paulo Gotardo, Thabo Beeler, Abhimitra Meka, and Kripasindhu Sarkar.
\newblock Cafca: High-quality novel view synthesis of expressive faces from casual few-shot captures.
\newblock In \emph{ACM SIGGRAPH Asia 2024 Conference Paper}, 2024.

\bibitem[B{\"u}hler et~al.(2023)B{\"u}hler, Sarkar, Shah, Li, Wang, Helminger, Orts-Escolano, Lagun, Hilliges, Beeler, et~al.]{buhler2023preface}
Marcel~C B{\"u}hler, Kripasindhu Sarkar, Tanmay Shah, Gengyan Li, Daoye Wang, Leonhard Helminger, Sergio Orts-Escolano, Dmitry Lagun, Otmar Hilliges, Thabo Beeler, et~al.
\newblock Preface: A data-driven volumetric prior for few-shot ultra high-resolution face synthesis.
\newblock In \emph{Proceedings of the IEEE/CVF International Conference on Computer Vision}, pages 3402--3413, 2023.

\bibitem[Cao et~al.(2022)Cao, Simon, Kim, Schwartz, Zollhoefer, Saito, Lombardi, Wei, Belko, Yu, Sheikh, and Saragih]{10.1145/3528223.3530143}
Chen Cao, Tomas Simon, Jin~Kyu Kim, Gabe Schwartz, Michael Zollhoefer, Shun-Suke Saito, Stephen Lombardi, Shih-En Wei, Danielle Belko, Shoou-I Yu, Yaser Sheikh, and Jason Saragih.
\newblock Authentic volumetric avatars from a phone scan.
\newblock \emph{ACM Trans. Graph.}, 41\penalty0 (4), 2022.

\bibitem[Chatziagapi et~al.(2024)Chatziagapi, Chrysos, and Samaras]{chatziagapi2024migs}
Aggelina Chatziagapi, Grigorios~G. Chrysos, and Dimitris Samaras.
\newblock Migs: Multi-identity gaussian splatting via tensor decomposition.
\newblock In \emph{ECCV}, 2024.

\bibitem[Chen et~al.(2023)Chen, Yi, Ma, Jia, and Lu]{Chen_2023_CVPR}
Jianchuan Chen, Wentao Yi, Liqian Ma, Xu Jia, and Huchuan Lu.
\newblock Gm-nerf: Learning generalizable model-based neural radiance fields from multi-view images.
\newblock In \emph{Proceedings of the IEEE/CVF Conference on Computer Vision and Pattern Recognition (CVPR)}, pages 20648--20658, 2023.

\bibitem[Chen et~al.(2022)Chen, Zhang, Xu, Liu, Cai, Feng, and Yan]{chen2022gpnerf}
Mingfei Chen, Jianfeng Zhang, Xiangyu Xu, Lijuan Liu, Yujun Cai, Jiashi Feng, and Shuicheng Yan.
\newblock Geometry-guided progressive nerf for generalizable and efficient neural human rendering.
\newblock In \emph{ECCV}, 2022.

\bibitem[Chen et~al.(2024{\natexlab{a}})Chen, Zheng, Li, Xu, and Liu]{chen2024meshavatar}
Yushuo Chen, Zerong Zheng, Zhe Li, Chao Xu, and Yebin Liu.
\newblock Meshavatar: Learning high-quality triangular human avatars from multi-view videos.
\newblock In \emph{ECCV}, 2024{\natexlab{a}}.

\bibitem[Chen et~al.(2024{\natexlab{b}})Chen, Moon, Guo, Cao, Pidhorskyi, Simon, Joshi, Dong, Xu, Pires, Wen, Evans, Peng, Buffalini, Trimble, McPhail, Schoeller, Yu, Romero, Zollhöfer, Sheikh, Liu, and Saito]{chen2024urhand}
Zhaoxi Chen, Gyeongsik Moon, Kaiwen Guo, Chen Cao, Stanislav Pidhorskyi, Tomas Simon, Rohan Joshi, Yuan Dong, Yichen Xu, Bernardo Pires, He Wen, Lucas Evans, Bo Peng, Julia Buffalini, Autumn Trimble, Kevyn McPhail, Melissa Schoeller, Shoou-I Yu, Javier Romero, Michael Zollhöfer, Yaser Sheikh, Ziwei Liu, and Shunsuke Saito.
\newblock {U}{R}hand: Universal relightable hands.
\newblock In \emph{CVPR}, 2024{\natexlab{b}}.

\bibitem[Corona et~al.(2022)Corona, Hodan, Vo, Moreno-Noguer, Sweeney, Newcombe, and Ma]{Corona_2022_CVPR}
Enric Corona, Tomas Hodan, Minh Vo, Francesc Moreno-Noguer, Chris Sweeney, Richard Newcombe, and Lingni Ma.
\newblock Lisa: Learning implicit shape and appearance of hands.
\newblock In \emph{Proceedings of the IEEE/CVF Conference on Computer Vision and Pattern Recognition (CVPR)}, pages 20533--20543, 2022.

\bibitem[Esser et~al.(2024)Esser, Kulal, Blattmann, Entezari, M{\"u}ller, Saini, Levi, Lorenz, Sauer, Boesel, et~al.]{esser2024scaling}
Patrick Esser, Sumith Kulal, Andreas Blattmann, Rahim Entezari, Jonas M{\"u}ller, Harry Saini, Yam Levi, Dominik Lorenz, Axel Sauer, Frederic Boesel, et~al.
\newblock Scaling rectified flow transformers for high-resolution image synthesis.
\newblock In \emph{Forty-first International Conference on Machine Learning}, 2024.

\bibitem[Feng et~al.(2023)Feng, Liu, Lai, Yang, and Li]{feng2023foflearningfourieroccupancy}
Qiao Feng, Yebin Liu, Yu-Kun Lai, Jingyu Yang, and Kun Li.
\newblock Fof: Learning fourier occupancy field for monocular real-time human reconstruction, 2023.

\bibitem[Guo et~al.(2021)Guo, Chen, Song, and Hilliges]{guo2021human}
Chen Guo, Xu Chen, Jie Song, and Otmar Hilliges.
\newblock Human performance capture from monocular video in the wild.
\newblock In \emph{2021 International Conference on 3D Vision (3DV)}, pages 889--898. IEEE, 2021.

\bibitem[Guo et~al.(2023)Guo, Jiang, Chen, Song, and Hilliges]{guo2023vid2avatar}
Chen Guo, Tianjian Jiang, Xu Chen, Jie Song, and Otmar Hilliges.
\newblock Vid2avatar: 3d avatar reconstruction from videos in the wild via self-supervised scene decomposition.
\newblock In \emph{Proceedings of the IEEE/CVF Conference on Computer Vision and Pattern Recognition (CVPR)}, 2023.

\bibitem[Guo et~al.(2024)Guo, Jiang, Kaufmann, Zheng, Valentin, Song, and Hilliges]{reloo}
Chen Guo, Tianjian Jiang, Manuel Kaufmann, Chengwei Zheng, Julien Valentin, Jie Song, and Otmar Hilliges.
\newblock Reloo: Reconstructing humans dressed in loose garments from monocular video in the wild.
\newblock In \emph{European conference on computer vision (ECCV)}, 2024.

\bibitem[Habermann et~al.(2020)Habermann, Xu, Zollhoefer, Pons-Moll, and Theobalt]{deepcap}
Marc Habermann, Weipeng Xu, Michael Zollhoefer, Gerard Pons-Moll, and Christian Theobalt.
\newblock Deepcap: Monocular human performance capture using weak supervision.
\newblock In \emph{{IEEE} Conference on Computer Vision and Pattern Recognition (CVPR)}. {IEEE}, 2020.

\bibitem[Habermann et~al.(2021)Habermann, Liu, Xu, Zollhoefer, Pons-Moll, and Theobalt]{habermann2021}
Marc Habermann, Lingjie Liu, Weipeng Xu, Michael Zollhoefer, Gerard Pons-Moll, and Christian Theobalt.
\newblock Real-time deep dynamic characters.
\newblock \emph{ACM Transactions on Graphics}, 40\penalty0 (4), 2021.

\bibitem[He et~al.(2021)He, Xu, Saito, Soatto, and Tung]{He_2021_ICCV}
Tong He, Yuanlu Xu, Shunsuke Saito, Stefano Soatto, and Tony Tung.
\newblock Arch++: Animation-ready clothed human reconstruction revisited.
\newblock In \emph{Proceedings of the IEEE/CVF International Conference on Computer Vision (ICCV)}, pages 11046--11056, 2021.

\bibitem[Ho et~al.(2024)Ho, Song, and Hilliges]{ho2024sith}
Hsuan-I Ho, Jie Song, and Otmar Hilliges.
\newblock Sith: Single-view textured human reconstruction with image-conditioned diffusion.
\newblock In \emph{Proceedings of the IEEE Conference on Computer Vision and Pattern Recognition (CVPR)}, 2024.

\bibitem[Hu et~al.(2024{\natexlab{a}})Hu, Zhang, Zhang, Zhou, Liu, Zhang, and Nie]{hu2024gaussianavatar}
Liangxiao Hu, Hongwen Zhang, Yuxiang Zhang, Boyao Zhou, Boning Liu, Shengping Zhang, and Liqiang Nie.
\newblock Gaussianavatar: Towards realistic human avatar modeling from a single video via animatable 3d gaussians.
\newblock In \emph{IEEE/CVF Conference on Computer Vision and Pattern Recognition (CVPR)}, 2024{\natexlab{a}}.

\bibitem[Hu et~al.(2024{\natexlab{b}})Hu, Hu, and Liu]{Hu_2024_CVPR}
Shoukang Hu, Tao Hu, and Ziwei Liu.
\newblock Gauhuman: Articulated gaussian splatting from monocular human videos.
\newblock In \emph{Proceedings of the IEEE/CVF Conference on Computer Vision and Pattern Recognition (CVPR)}, pages 20418--20431, 2024{\natexlab{b}}.

\bibitem[Hu et~al.(2022)Hu, Yu, Zheng, Zhang, Liu, and Zwicker]{HVTR:3DV2022}
Tao Hu, Tao Yu, Zerong Zheng, He Zhang, Yebin Liu, and Matthias Zwicker.
\newblock Hvtr: Hybrid volumetric-textural rendering for human avatars.
\newblock In \emph{2022 International Conference on 3D Vision (3DV)}, 2022.

\bibitem[Huang et~al.(2024)Huang, Yi, Xiu, Liao, Tang, Cai, and Thies]{huang2024tech}
Yangyi Huang, Hongwei Yi, Yuliang Xiu, Tingting Liao, Jiaxiang Tang, Deng Cai, and Justus Thies.
\newblock {TeCH: Text-guided Reconstruction of Lifelike Clothed Humans}.
\newblock In \emph{International Conference on 3D Vision (3DV)}, 2024.

\bibitem[Huang et~al.(2020)Huang, Xu, Lassner, Li, and Tung]{huang2020arch}
Zeng Huang, Yuanlu Xu, Christoph Lassner, Hao Li, and Tony Tung.
\newblock Arch: Animatable reconstruction of clothed humans.
\newblock In \emph{Proceedings of the IEEE/CVF Conference on Computer Vision and Pattern Recognition}, pages 3093--3102, 2020.

\bibitem[Jiakai et~al.(2021)Jiakai, Xinhang, Xinyi, Fuqiang, Yanshun, Minye, Yingliang, Lan, and Jingyi]{zhang2021stnerf}
Zhang Jiakai, Liu Xinhang, Ye Xinyi, Zhao Fuqiang, Zhang Yanshun, Wu Minye, Zhang Yingliang, Xu Lan, and Yu Jingyi.
\newblock Editable free-viewpoint video using a layered neural representation.
\newblock In \emph{ACM SIGGRAPH}, 2021.

\bibitem[Jiang et~al.(2022{\natexlab{a}})Jiang, Hong, Bao, and Zhang]{jiang2022selfrecon}
Boyi Jiang, Yang Hong, Hujun Bao, and Juyong Zhang.
\newblock Selfrecon: Self reconstruction your digital avatar from monocular video.
\newblock In \emph{{IEEE/CVF} Conference on Computer Vision and Pattern Recognition (CVPR)}, 2022{\natexlab{a}}.

\bibitem[Jiang et~al.(2023)Jiang, Chen, Song, and Hilliges]{jiang2023instantavatar}
Tianjian Jiang, Xu Chen, Jie Song, and Otmar Hilliges.
\newblock Instantavatar: Learning avatars from monocular video in 60 seconds.
\newblock In \emph{Proceedings of the IEEE/CVF Conference on Computer Vision and Pattern Recognition (CVPR)}, 2023.

\bibitem[Jiang et~al.(2022{\natexlab{b}})Jiang, Yi, Samei, Tuzel, and Ranjan]{jiang2022neuman}
Wei Jiang, Kwang~Moo Yi, Golnoosh Samei, Oncel Tuzel, and Anurag Ranjan.
\newblock Neuman: Neural human radiance field from a single video.
\newblock In \emph{Proceedings of the European conference on computer vision (ECCV)}, 2022{\natexlab{b}}.

\bibitem[Jiang et~al.(2024)Jiang, Guo, Kaufmann, Jiang, Valentin, Hilliges, and Song]{multiply}
Zeren Jiang, Chen Guo, Manuel Kaufmann, Tianjian Jiang, Julien Valentin, Otmar Hilliges, and Jie Song.
\newblock Multiply: Reconstruction of multiple people from monocular video in the wild.
\newblock In \emph{Proceedings of the IEEE/CVF Conference on Computer Vision and Pattern Recognition (CVPR)}, 2024.

\bibitem[Joo et~al.(2018)Joo, Simon, and Sheikh]{Joo_2018_CVPR}
Hanbyul Joo, Tomas Simon, and Yaser Sheikh.
\newblock Total capture: A 3d deformation model for tracking faces, hands, and bodies.
\newblock In \emph{Proceedings of the IEEE Conference on Computer Vision and Pattern Recognition (CVPR)}, 2018.

\bibitem[Jung et~al.(2023)Jung, Brasch, Song, Perez-Pellitero, Zhou, Li, Navab, and Busam]{jung2023deformable3dgaussiansplatting}
HyunJun Jung, Nikolas Brasch, Jifei Song, Eduardo Perez-Pellitero, Yiren Zhou, Zhihao Li, Nassir Navab, and Benjamin Busam.
\newblock Deformable 3d gaussian splatting for animatable human avatars, 2023.

\bibitem[Ke et~al.(2023)Ke, Ye, Danelljan, Liu, Tai, Tang, and Yu]{sam_hq}
Lei Ke, Mingqiao Ye, Martin Danelljan, Yifan Liu, Yu-Wing Tai, Chi-Keung Tang, and Fisher Yu.
\newblock Segment anything in high quality.
\newblock In \emph{NeurIPS}, 2023.

\bibitem[Kerbl et~al.(2023)Kerbl, Kopanas, Leimk{\"u}hler, and Drettakis]{kerbl3Dgaussians}
Bernhard Kerbl, Georgios Kopanas, Thomas Leimk{\"u}hler, and George Drettakis.
\newblock 3d gaussian splatting for real-time radiance field rendering.
\newblock \emph{ACM Transactions on Graphics}, 42\penalty0 (4), 2023.

\bibitem[Khirodkar et~al.(2025)Khirodkar, Bagautdinov, Martinez, Zhaoen, James, Selednik, Anderson, and Saito]{10.1007/978-3-031-73235-5_12}
Rawal Khirodkar, Timur Bagautdinov, Julieta Martinez, Su Zhaoen, Austin James, Peter Selednik, Stuart Anderson, and Shunsuke Saito.
\newblock Sapiens: Foundation for human vision models.
\newblock In \emph{Computer Vision -- ECCV 2024}, pages 206--228, Cham, 2025. Springer Nature Switzerland.

\bibitem[Kirillov et~al.(2023)Kirillov, Mintun, Ravi, Mao, Rolland, Gustafson, Xiao, Whitehead, Berg, Lo, Dollar, and Girshick]{Kirillov_2023_ICCV}
Alexander Kirillov, Eric Mintun, Nikhila Ravi, Hanzi Mao, Chloe Rolland, Laura Gustafson, Tete Xiao, Spencer Whitehead, Alexander~C. Berg, Wan-Yen Lo, Piotr Dollar, and Ross Girshick.
\newblock Segment anything.
\newblock In \emph{Proceedings of the IEEE/CVF International Conference on Computer Vision (ICCV)}, pages 4015--4026, 2023.

\bibitem[Kocabas et~al.(2024)Kocabas, Chang, Gabriel, Tuzel, and Ranjan]{kocabas2024hugs}
Muhammed Kocabas, Jen-Hao~Rick Chang, James Gabriel, Oncel Tuzel, and Anurag Ranjan.
\newblock {HUGS}: Human gaussian splatting.
\newblock In \emph{2024 IEEE/CVF Conference on Computer Vision and Pattern Recognition (CVPR)}, 2024.

\bibitem[Kwon et~al.(2021)Kwon, Kim, Ceylan, and Fuchs]{kwon2021neural}
Youngjoong Kwon, Dahun Kim, Duygu Ceylan, and Henry Fuchs.
\newblock Neural human performer: Learning generalizable radiance fields for human performance rendering.
\newblock In \emph{Advances in Neural Information Processing Systems}, 2021.

\bibitem[Kwon et~al.(2024)Kwon, Fang, Lu, Dong, Zhang, Carrasco, Mosella-Montoro, Xu, Takagi, Kim, Prakash, and la~Torre]{kwon2024ghg}
Youngjoong Kwon, Baole Fang, Yixing Lu, Haoye Dong, Cheng Zhang, Francisco~Vicente Carrasco, Albert Mosella-Montoro, Jianjin Xu, Shingo Takagi, Daeil Kim, Aayush Prakash, and Fernando~De la Torre.
\newblock Generalizable human gaussians for sparse view synthesis.
\newblock In \emph{European Conference on Computer Vision}, 2024.

\bibitem[Lei et~al.(2024)Lei, Wang, Pavlakos, Liu, and Daniilidis]{Lei_2024_CVPR}
Jiahui Lei, Yufu Wang, Georgios Pavlakos, Lingjie Liu, and Kostas Daniilidis.
\newblock Gart: Gaussian articulated template models.
\newblock In \emph{Proceedings of the IEEE/CVF Conference on Computer Vision and Pattern Recognition (CVPR)}, pages 19876--19887, 2024.

\bibitem[Li et~al.(2024{\natexlab{a}})Li, Cao, Schwartz, Khirodkar, Richardt, Simon, Sheikh, and Saito]{li2024uravatar}
Junxuan Li, Chen Cao, Gabriel Schwartz, Rawal Khirodkar, Christian Richardt, Tomas Simon, Yaser Sheikh, and Shunsuke Saito.
\newblock Uravatar: Universal relightable gaussian codec avatars.
\newblock In \emph{ACM SIGGRAPH 2024 Conference Papers}, 2024{\natexlab{a}}.

\bibitem[Li et~al.(2023{\natexlab{a}})Li, Tao, Yang, and Yang]{li2023human101}
Mingwei Li, Jiachen Tao, Zongxin Yang, and Yi Yang.
\newblock Human101: Training 100+fps human gaussians in 100s from 1 view, 2023{\natexlab{a}}.

\bibitem[Li et~al.(2024{\natexlab{b}})Li, Yao, Xie, and Chen]{li2024gaussianbodyclothedhumanreconstruction}
Mengtian Li, Shengxiang Yao, Zhifeng Xie, and Keyu Chen.
\newblock Gaussianbody: Clothed human reconstruction via 3d gaussian splatting, 2024{\natexlab{b}}.

\bibitem[Li et~al.(2022)Li, Tanke, Vo, Zollhofer, Gall, Kanazawa, and Lassner]{li2022tava}
Ruilong Li, Julian Tanke, Minh Vo, Michael Zollhofer, Jurgen Gall, Angjoo Kanazawa, and Christoph Lassner.
\newblock Tava: Template-free animatable volumetric actors.
\newblock In \emph{European Conference on Computer Vision (ECCV)}, 2022.

\bibitem[Li et~al.(2023{\natexlab{b}})Li, Zheng, Liu, Zhou, and Liu]{li2023posevocab}
Zhe Li, Zerong Zheng, Yuxiao Liu, Boyao Zhou, and Yebin Liu.
\newblock Posevocab: Learning joint-structured pose embeddings for human avatar modeling.
\newblock \emph{ACM SIGGRAPH Conference Proceedings}, 2023{\natexlab{b}}.

\bibitem[Li et~al.(2024{\natexlab{c}})Li, Zheng, Wang, and Liu]{li2024animatablegaussians}
Zhe Li, Zerong Zheng, Lizhen Wang, and Yebin Liu.
\newblock Animatable gaussians: Learning pose-dependent gaussian maps for high-fidelity human avatar modeling.
\newblock In \emph{Proceedings of the IEEE/CVF Conference on Computer Vision and Pattern Recognition (CVPR)}, 2024{\natexlab{c}}.

\bibitem[Lin et~al.(2022)Lin, Zhang, Zheng, Shao, and Liu]{lin2022fite}
Siyou Lin, Hongwen Zhang, Zerong Zheng, Ruizhi Shao, and Yebin Liu.
\newblock Learning implicit templates for point-based clothed human modeling.
\newblock In \emph{ECCV}, 2022.

\bibitem[Liu et~al.(2021)Liu, Habermann, Rudnev, Sarkar, Gu, and Theobalt]{liu2021neural}
Lingjie Liu, Marc Habermann, Viktor Rudnev, Kripasindhu Sarkar, Jiatao Gu, and Christian Theobalt.
\newblock Neural actor: Neural free-view synthesis of human actors with pose control.
\newblock \emph{ACM Trans. Graph.(ACM SIGGRAPH Asia)}, 2021.

\bibitem[Liu et~al.(2024)Liu, Wu, Liu, Liu, Wu, Zhao, Feng, Ding, and Wang]{liu2024gvareconstructingvivid3d}
Xinqi Liu, Chenming Wu, Jialun Liu, Xing Liu, Jinbo Wu, Chen Zhao, Haocheng Feng, Errui Ding, and Jingdong Wang.
\newblock Gva: Reconstructing vivid 3d gaussian avatars from monocular videos, 2024.

\bibitem[Loper et~al.(2015)Loper, Mahmood, Romero, Pons-Moll, and Black]{loper2015smpl}
Matthew Loper, Naureen Mahmood, Javier Romero, Gerard Pons-Moll, and Michael~J Black.
\newblock Smpl: A skinned multi-person linear model.
\newblock \emph{ACM transactions on graphics (TOG)}, 34\penalty0 (6):\penalty0 1--16, 2015.

\bibitem[Mildenhall et~al.(2020)Mildenhall, Srinivasan, Tancik, Barron, Ramamoorthi, and Ng]{mildenhall2020nerf}
Ben Mildenhall, Pratul~P Srinivasan, Matthew Tancik, Jonathan~T Barron, Ravi Ramamoorthi, and Ren Ng.
\newblock Nerf: Representing scenes as neural radiance fields for view synthesis.
\newblock In \emph{European conference on computer vision}, pages 405--421. Springer, 2020.

\bibitem[Moon et~al.(2024{\natexlab{a}})Moon, Shiratori, and Saito]{moon2024exavatar}
Gyeongsik Moon, Takaaki Shiratori, and Shunsuke Saito.
\newblock Expressive whole-body {3D} gaussian avatar.
\newblock In \emph{ECCV}, 2024{\natexlab{a}}.

\bibitem[Moon et~al.(2024{\natexlab{b}})Moon, Xu, Joshi, Wu, and Shiratori]{Moon_2024_CVPR}
Gyeongsik Moon, Weipeng Xu, Rohan Joshi, Chenglei Wu, and Takaaki Shiratori.
\newblock Authentic hand avatar from a phone scan via universal hand model.
\newblock In \emph{Proceedings of the IEEE/CVF Conference on Computer Vision and Pattern Recognition (CVPR)}, pages 2029--2038, 2024{\natexlab{b}}.

\bibitem[Moreau et~al.(2024)Moreau, Song, Dhamo, Shaw, Zhou, and P{\'e}rez-Pellitero]{moreau2024human}
Arthur Moreau, Jifei Song, Helisa Dhamo, Richard Shaw, Yiren Zhou, and Eduardo P{\'e}rez-Pellitero.
\newblock Human gaussian splatting: Real-time rendering of animatable avatars.
\newblock In \emph{CVPR}, 2024.

\bibitem[Noguchi et~al.(2021)Noguchi, Sun, Lin, and Harada]{2021narf}
Atsuhiro Noguchi, Xiao Sun, Stephen Lin, and Tatsuya Harada.
\newblock Neural articulated radiance field.
\newblock In \emph{International Conference on Computer Vision}, 2021.

\bibitem[Pang et~al.(2024)Pang, Zhu, Kortylewski, Theobalt, and Habermann]{Pang_2024_CVPR}
Haokai Pang, Heming Zhu, Adam Kortylewski, Christian Theobalt, and Marc Habermann.
\newblock Ash: Animatable gaussian splats for efficient and photoreal human rendering.
\newblock In \emph{Proceedings of the IEEE/CVF Conference on Computer Vision and Pattern Recognition (CVPR)}, pages 1165--1175, 2024.

\bibitem[Pavlakos et~al.(2019)Pavlakos, Choutas, Ghorbani, Bolkart, Osman, Tzionas, and Black]{SMPL-X:2019}
Georgios Pavlakos, Vasileios Choutas, Nima Ghorbani, Timo Bolkart, Ahmed A.~A. Osman, Dimitrios Tzionas, and Michael~J. Black.
\newblock Expressive body capture: {3D} hands, face, and body from a single image.
\newblock In \emph{Proceedings IEEE Conf. on Computer Vision and Pattern Recognition (CVPR)}, pages 10975--10985, 2019.

\bibitem[Peebles and Xie(2023)]{peebles2023scalable}
William Peebles and Saining Xie.
\newblock Scalable diffusion models with transformers.
\newblock In \emph{Proceedings of the IEEE/CVF International Conference on Computer Vision}, pages 4195--4205, 2023.

\bibitem[Peng et~al.(2021{\natexlab{a}})Peng, Dong, Wang, Zhang, Shuai, Zhou, and Bao]{peng2021animatable}
Sida Peng, Junting Dong, Qianqian Wang, Shangzhan Zhang, Qing Shuai, Xiaowei Zhou, and Hujun Bao.
\newblock Animatable neural radiance fields for modeling dynamic human bodies.
\newblock In \emph{ICCV}, 2021{\natexlab{a}}.

\bibitem[Peng et~al.(2021{\natexlab{b}})Peng, Zhang, Xu, Wang, Shuai, Bao, and Zhou]{peng2021neural}
Sida Peng, Yuanqing Zhang, Yinghao Xu, Qianqian Wang, Qing Shuai, Hujun Bao, and Xiaowei Zhou.
\newblock Neural body: Implicit neural representations with structured latent codes for novel view synthesis of dynamic humans.
\newblock In \emph{Proceedings of the IEEE/CVF Conference on Computer Vision and Pattern Recognition}, pages 9054--9063, 2021{\natexlab{b}}.

\bibitem[Qian et~al.(2024)Qian, Wang, Mihajlovic, Geiger, and Tang]{qian20233dgsavatar}
Zhiyin Qian, Shaofei Wang, Marko Mihajlovic, Andreas Geiger, and Siyu Tang.
\newblock 3dgs-avatar: Animatable avatars via deformable 3d gaussian splatting.
\newblock In \emph{IEEE/CVF Conference on Computer Vision and Pattern Recognition (CVPR)}, 2024.

\bibitem[Remelli et~al.(2022)Remelli, Bagautdinov, Saito, Wu, Simon, Wei, Guo, Cao, Prada, Saragih, et~al.]{remelli2022drivable}
Edoardo Remelli, Timur Bagautdinov, Shunsuke Saito, Chenglei Wu, Tomas Simon, Shih-En Wei, Kaiwen Guo, Zhe Cao, Fabian Prada, Jason Saragih, et~al.
\newblock Drivable volumetric avatars using texel-aligned features.
\newblock In \emph{ACM SIGGRAPH 2022 Conference Proceedings}, pages 1--9, 2022.

\bibitem[Rong et~al.(2021)Rong, Shiratori, and Joo]{rong2021frankmocap}
Yu Rong, Takaaki Shiratori, and Hanbyul Joo.
\newblock Frankmocap: A monocular 3d whole-body pose estimation system via regression and integration.
\newblock In \emph{IEEE International Conference on Computer Vision Workshops}, 2021.

\bibitem[Ronneberger et~al.(2015)Ronneberger, Fischer, and Brox]{unet}
Olaf Ronneberger, Philipp Fischer, and Thomas Brox.
\newblock U-net: Convolutional networks for biomedical image segmentation.
\newblock In \emph{Medical Image Computing and Computer-Assisted Intervention -- MICCAI 2015}, pages 234--241, Cham, 2015. Springer International Publishing.

\bibitem[Saito et~al.(2019)Saito, Huang, Natsume, Morishima, Kanazawa, and Li]{saito2019pifu}
Shunsuke Saito, Zeng Huang, Ryota Natsume, Shigeo Morishima, Angjoo Kanazawa, and Hao Li.
\newblock Pifu: Pixel-aligned implicit function for high-resolution clothed human digitization.
\newblock In \emph{Proceedings of the IEEE/CVF International Conference on Computer Vision}, pages 2304--2314, 2019.

\bibitem[Saito et~al.(2020)Saito, Simon, Saragih, and Joo]{saito2020pifuhd}
Shunsuke Saito, Tomas Simon, Jason Saragih, and Hanbyul Joo.
\newblock Pifuhd: Multi-level pixel-aligned implicit function for high-resolution 3d human digitization.
\newblock In \emph{Proceedings of the IEEE/CVF Conference on Computer Vision and Pattern Recognition}, pages 84--93, 2020.

\bibitem[Saito et~al.(2024)Saito, Schwartz, Simon, Li, and Nam]{saito2024rgca}
Shunsuke Saito, Gabriel Schwartz, Tomas Simon, Junxuan Li, and Giljoo Nam.
\newblock Relightable gaussian codec avatars.
\newblock In \emph{CVPR}, 2024.

\bibitem[Shao et~al.(2024)Shao, Wang, Li, Wang, Lin, Zhang, Fan, and Wang]{shao2024splattingavatar}
Zhijing Shao, Zhaolong Wang, Zhuang Li, Duotun Wang, Xiangru Lin, Yu Zhang, Mingming Fan, and Zeyu Wang.
\newblock {SplattingAvatar: Realistic Real-Time Human Avatars with Mesh-Embedded Gaussian Splatting}.
\newblock In \emph{Proceedings of the IEEE/CVF Conference on Computer Vision and Pattern Recognition (CVPR)}, 2024.

\bibitem[Shen et~al.(2023)Shen, Guo, Kaufmann, Zarate, Valentin, Song, and Hilliges]{shen2023xavatar}
Kaiyue Shen, Chen Guo, Manuel Kaufmann, Juan Zarate, Julien Valentin, Jie Song, and Otmar Hilliges.
\newblock X-avatar: Expressive human avatars.
\newblock In \emph{Computer Vision and Pattern Recognition (CVPR)}, 2023.

\bibitem[Su et~al.(2021)Su, Yu, Zollh{\"o}fer, and Rhodin]{su2021anerf}
Shih-Yang Su, Frank Yu, Michael Zollh{\"o}fer, and Helge Rhodin.
\newblock A-nerf: Articulated neural radiance fields for learning human shape, appearance, and pose.
\newblock In \emph{Advances in Neural Information Processing Systems}, 2021.

\bibitem[Su et~al.(2022)Su, Bagautdinov, and Rhodin]{su2022danbo}
Shih-Yang Su, Timur Bagautdinov, and Helge Rhodin.
\newblock Danbo: Disentangled articulated neural body representations via graph neural networks.
\newblock In \emph{European Conference on Computer Vision}, 2022.

\bibitem[Sun et~al.(2024{\natexlab{a}})Sun, Dabral, Fua, Theobalt, and Habermann]{sun2024metacap}
Guoxing Sun, Rishabh Dabral, Pascal Fua, Christian Theobalt, and Marc Habermann.
\newblock Metacap: Meta-learning priors from multi-view imagery for sparse-view human performance capture and rendering.
\newblock In \emph{ECCV}, 2024{\natexlab{a}}.

\bibitem[Sun et~al.(2024{\natexlab{b}})Sun, Wang, Zeng, Yin, Wei, Wang, Mei, Leung, Liu, Yang, and Cai]{Sun_2024_CVPR}
Qingping Sun, Yanjun Wang, Ailing Zeng, Wanqi Yin, Chen Wei, Wenjia Wang, Haiyi Mei, Chi-Sing Leung, Ziwei Liu, Lei Yang, and Zhongang Cai.
\newblock Aios: All-in-one-stage expressive human pose and shape estimation.
\newblock In \emph{Proceedings of the IEEE/CVF Conference on Computer Vision and Pattern Recognition (CVPR)}, pages 1834--1843, 2024{\natexlab{b}}.

\bibitem[Svitov et~al.(2024)Svitov, Morerio, Agapito, and Bue]{svitov2024hahahighlyarticulatedgaussian}
David Svitov, Pietro Morerio, Lourdes Agapito, and Alessio~Del Bue.
\newblock Haha: Highly articulated gaussian human avatars with textured mesh prior, 2024.

\bibitem[Wang et~al.(2022)Wang, Schwarz, Geiger, and Tang]{ARAH:ECCV:2022}
Shaofei Wang, Katja Schwarz, Andreas Geiger, and Siyu Tang.
\newblock Arah: Animatable volume rendering of articulated human sdfs.
\newblock In \emph{European Conference on Computer Vision (ECCV)}, 2022.

\bibitem[Wang et~al.(2003)Wang, Simoncelli, and Bovik]{wang2003multiscale}
Zhou Wang, Eero~P Simoncelli, and Alan~C Bovik.
\newblock Multiscale structural similarity for image quality assessment.
\newblock In \emph{The Thrity-Seventh Asilomar Conference on Signals, Systems \& Computers}, 2003.

\bibitem[Wen et~al.(2024)Wen, Zhao, Ren, Schwing, and Wang]{wen2024gomavatar}
Jing Wen, Xiaoming Zhao, Zhongzheng Ren, Alex Schwing, and Shenlong Wang.
\newblock {GoMAvatar: Efficient Animatable Human Modeling from Monocular Video Using Gaussians-on-Mesh}.
\newblock In \emph{CVPR}, 2024.

\bibitem[Weng et~al.(2022)Weng, Curless, Srinivasan, Barron, and Kemelmacher-Shlizerman]{weng_humannerf_2022_cvpr}
Chung-Yi Weng, Brian Curless, Pratul~P. Srinivasan, Jonathan~T. Barron, and Ira Kemelmacher-Shlizerman.
\newblock Human{N}e{RF}: Free-viewpoint rendering of moving people from monocular video.
\newblock In \emph{Proceedings of the IEEE/CVF Conference on Computer Vision and Pattern Recognition (CVPR)}, pages 16210--16220, 2022.

\bibitem[Xiang et~al.(2021)Xiang, Prada, Bagautdinov, Xu, Dong, Wen, Hodgins, and Wu]{10.1145/3478513.3480545}
Donglai Xiang, Fabian Prada, Timur Bagautdinov, Weipeng Xu, Yuan Dong, He Wen, Jessica Hodgins, and Chenglei Wu.
\newblock Modeling clothing as a separate layer for an animatable human avatar.
\newblock \emph{ACM Trans. Graph.}, 40\penalty0 (6), 2021.

\bibitem[Xiu et~al.(2022)Xiu, Yang, Tzionas, and Black]{xiu2022icon}
Yuliang Xiu, Jinlong Yang, Dimitrios Tzionas, and Michael~J. Black.
\newblock {ICON}: {I}mplicit {C}lothed humans {O}btained from {N}ormals.
\newblock In \emph{Proceedings of the IEEE/CVF Conference on Computer Vision and Pattern Recognition (CVPR)}, pages 13296--13306, 2022.

\bibitem[Xiu et~al.(2023)Xiu, Yang, Cao, Tzionas, and Black]{xiu2023econ}
Yuliang Xiu, Jinlong Yang, Xu Cao, Dimitrios Tzionas, and Michael~J. Black.
\newblock {ECON: Explicit Clothed humans Optimized via Normal integration}.
\newblock In \emph{Proceedings of the IEEE/CVF Conference on Computer Vision and Pattern Recognition (CVPR)}, 2023.

\bibitem[Xiu et~al.(2024)Xiu, Ye, Liu, Tzionas, and Black]{xiu2024puzzleavatar}
Yuliang Xiu, Yufei Ye, Zhen Liu, Dimitrios Tzionas, and Michael~J Black.
\newblock Puzzleavatar: Assembling 3d avatars from personal albums.
\newblock \emph{ACM Transactions on Graphics (TOG)}, 2024.

\bibitem[Xu et~al.(2022)Xu, Fujita, and Matsumoto]{xu2022sanerf}
Tianhan Xu, Yasuhiro Fujita, and Eiichi Matsumoto.
\newblock Surface-aligned neural radiance fields for controllable 3d human synthesis.
\newblock In \emph{CVPR}, 2022.

\bibitem[Xu et~al.(2018)Xu, Chatterjee, Zollh\"{o}fer, Rhodin, Mehta, Seidel, and Theobalt]{Xu:2018:MHP:3191713.3181973}
Weipeng Xu, Avishek Chatterjee, Michael Zollh\"{o}fer, Helge Rhodin, Dushyant Mehta, Hans-Peter Seidel, and Christian Theobalt.
\newblock Monoperfcap: Human performance capture from monocular video.
\newblock \emph{SIGGRAPH}, 37\penalty0 (2):\penalty0 27:1--27:15, 2018.

\bibitem[Xue et~al.(2024)Xue, Guo, Zheng, Wang, Jiang, Ho, Kaufmann, Song, and Otmar]{xue2024hsr}
Lixin Xue, Chen Guo, Chengwei Zheng, Fangjinhua Wang, Tianjian Jiang, Hsuan-I Ho, Manuel Kaufmann, Jie Song, and Hilliges Otmar.
\newblock {HSR:} holistic 3d human-scene reconstruction from monocular videos.
\newblock In \emph{European Conference on Computer Vision (ECCV)}, 2024.

\bibitem[Yariv et~al.(2021)Yariv, Gu, Kasten, and Lipman]{yariv2021volume}
Lior Yariv, Jiatao Gu, Yoni Kasten, and Yaron Lipman.
\newblock Volume rendering of neural implicit surfaces.
\newblock In \emph{Advances in Neural Information Processing Systems}, 2021.

\bibitem[Yin et~al.(2023)Yin, Guo, Kaufmann, Zarate, Song, and Hilliges]{yin2023hi4d}
Yifei Yin, Chen Guo, Manuel Kaufmann, Juan Zarate, Jie Song, and Otmar Hilliges.
\newblock Hi4d: 4d instance segmentation of close human interaction.
\newblock In \emph{Computer Vision and Pattern Recognition (CVPR)}, 2023.

\bibitem[Yu et~al.(2023)Yu, Cheng, Liu, Wu, and Lin]{yu2023monohuman}
Zhengming Yu, Wei Cheng, Xian Liu, Wayne Wu, and Kwan-Yee Lin.
\newblock Monohuman: Animatable human neural field from monocular video.
\newblock In \emph{Proceedings of the IEEE/CVF Conference on Computer Vision and Pattern Recognition}, pages 16943--16953, 2023.

\bibitem[Zhang et~al.(2018)Zhang, Isola, Efros, Shechtman, and Wang]{zhang2018perceptual}
Richard Zhang, Phillip Isola, Alexei~A Efros, Eli Shechtman, and Oliver Wang.
\newblock The unreasonable effectiveness of deep features as a perceptual metric.
\newblock In \emph{CVPR}, 2018.

\bibitem[Zhao et~al.(2022)Zhao, Yang, Zhang, Lin, Zhang, Yu, and Xu]{Zhao_2022_CVPR}
Fuqiang Zhao, Wei Yang, Jiakai Zhang, Pei Lin, Yingliang Zhang, Jingyi Yu, and Lan Xu.
\newblock Humannerf: Efficiently generated human radiance field from sparse inputs.
\newblock In \emph{Proceedings of the IEEE/CVF Conference on Computer Vision and Pattern Recognition (CVPR)}, pages 7743--7753, 2022.

\bibitem[Zheng et~al.(2024{\natexlab{a}})Zheng, Zhou, Shao, Liu, Zhang, Nie, and Liu]{zheng2024gpsgaussian}
Shunyuan Zheng, Boyao Zhou, Ruizhi Shao, Boning Liu, Shengping Zhang, Liqiang Nie, and Yebin Liu.
\newblock Gps-gaussian: Generalizable pixel-wise 3d gaussian splatting for real-time human novel view synthesis.
\newblock In \emph{Proceedings of the IEEE/CVF Conference on Computer Vision and Pattern Recognition (CVPR)}, 2024{\natexlab{a}}.

\bibitem[Zheng et~al.(2024{\natexlab{b}})Zheng, Zhao, Yang, Yifan, Xiang, Dubost, Lagun, Beeler, Tombari, Guibas, and Wetzstein]{PhysAavatar24}
Yang Zheng, Qingqing Zhao, Guandao Yang, Wang Yifan, Donglai Xiang, Florian Dubost, Dmitry Lagun, Thabo Beeler, Federico Tombari, Leonidas Guibas, and Gordon Wetzstein.
\newblock Physavatar: Learning the physics of dressed 3d avatars from visual observations.
\newblock In \emph{European Conference on Computer Vision (ECCV)}, 2024{\natexlab{b}}.

\bibitem[Zheng et~al.(2021)Zheng, Yu, Liu, and Dai]{zheng2021pamir}
Zerong Zheng, Tao Yu, Yebin Liu, and Qionghai Dai.
\newblock Pamir: Parametric model-conditioned implicit representation for image-based human reconstruction.
\newblock \emph{IEEE Transactions on Pattern Analysis and Machine Intelligence}, 2021.

\bibitem[Zheng et~al.(2023)Zheng, Zhao, Zhang, Liu, and Liu]{zheng2023avatarrex}
Zerong Zheng, Xiaochen Zhao, Hongwen Zhang, Boning Liu, and Yebin Liu.
\newblock Avatarrex: Real-time expressive full-body avatars.
\newblock \emph{ACM Transactions on Graphics (TOG)}, 42\penalty0 (4), 2023.

\bibitem[Zhu et~al.(2024)Zhu, Zhan, Theobalt, and Habermann]{10.1145/3697140}
Heming Zhu, Fangneng Zhan, Christian Theobalt, and Marc Habermann.
\newblock Trihuman: A real-time and controllable tri-plane representation for detailed human geometry and appearance synthesis.
\newblock \emph{ACM Trans. Graph.}, 2024.

\bibitem[Zielonka et~al.(2023)Zielonka, Bagautdinov, Saito, Zollhöfer, Thies, and Romero]{zielonka2023drivable3dgaussianavatars}
Wojciech Zielonka, Timur Bagautdinov, Shunsuke Saito, Michael Zollhöfer, Justus Thies, and Javier Romero.
\newblock Drivable 3d gaussian avatars, 2023.

\end{thebibliography}
}

% WARNING: do not forget to delete the supplementary pages from your submission 
% \input{sec/X_suppl}

\end{document}